\documentclass[conference]{IEEEtran}
\IEEEoverridecommandlockouts
\usepackage{cite}
\usepackage{amsmath,amssymb,amsfonts}
\usepackage{algorithmic}
\usepackage{graphicx}
\usepackage{textcomp}
\usepackage{xcolor}
\def\BibTeX{{\rm B\kern-.05em{\sc i\kern-.025em b}\kern-.08em
    T\kern-.1667em\lower.7ex\hbox{E}\kern-.125emX}}

\usepackage{subcaption}
\usepackage{url}

\usepackage{booktabs}
\usepackage{multirow}

\DeclareMathOperator{\E}{\mathbb{E}}
\DeclareMathOperator{\s}{\mathbf{s}}
\DeclareMathOperator{\obs}{\mathbf{o}}
\DeclareMathOperator{\act}{\mathbf{a}}
\DeclareMathOperator{\h}{\mathbf{h}}
\DeclareMathOperator{\x}{\mathbf{x}}

\usepackage{algorithm}
\usepackage{algorithmic}

\usepackage{xcolor}
\definecolor{codegreen}{rgb}{0,0.6,0}
\definecolor{codegray}{rgb}{0.5,0.5,0.5}
\usepackage{newfloat}
\usepackage{listings}
\floatstyle{ruled}
\newfloat{listing}{tb}{lst}{}
\floatname{listing}{Listing}

\begin{document}

\title{Learning of feature points without additional supervision improves RL from images}

\author{
% \IEEEauthorblockN{Anonymous Authors}
% \IEEEauthorblockA{\textit{Department} \\
% \textit{Institution}\\
% email}
\IEEEauthorblockN{Rinu Boney, Alexander Ilin, Juho Kannala}
\IEEEauthorblockA{
% \textit{Department of Computer Science} \\
\textit{Aalto University, Finland}\\
% Espoo, Finland \\
% firstname.lastname@aalto.fi
}
% \and
% \IEEEauthorblockN{Alexander Ilin}
% \IEEEauthorblockA{\textit{Department of Computer Science} \\
% \textit{Aalto University}\\
% Espoo, Finland \\
% alexander.ilin@aalto.fi}
% \and
% \IEEEauthorblockN{Juho Kannala}
% \IEEEauthorblockA{\textit{Department of Computer Science} \\
% \textit{Aalto University}\\
% Espoo, Finland \\
% juho.kannala@aalto.fi}
}

\maketitle

\begin{abstract}
In many control problems that include vision, optimal controls can be inferred from the location of the objects in the scene. This information can be represented using feature points, which is a list of spatial locations in learned feature maps of an input image. Previous works show that feature points learned using unsupervised pre-training or human supervision can provide good features for control tasks. In this paper, we show that it is possible to learn efficient feature point representations end-to-end, without the need for unsupervised pre-training, decoders, or additional losses. Our proposed architecture consists of a differentiable feature point extractor that feeds the coordinates of the estimated feature points directly to a soft actor-critic agent. The proposed algorithm yields performance competitive to the state-of-the art on DeepMind Control Suite tasks.
\end{abstract}


%===============================================================================

\section{Introduction}
	
% Motivation SRL
Learning state representations of image observations for control is highly challenging as images are high-dimensional and supervisory signals in reinforcement learning (RL) are limited to scalar rewards. %State representation learning for continuous control from images is an area of active research.
State-of-the-art model-free RL algorithms with convolutional encoders, without any additional enhancements, have been highly data-inefficient in this setting~\cite{tassa2020dmcontrol, yarats2019improving}.
A popular solution is to enhance the supervision of the encoder by introducing an auxiliary autoencoding task%
%Autoencoders are popular unsupervised representation learning methods and they have been successfully used to improve state representation learning for continuous control from images~\cite{yarats2019improving, lange2010deep, shelhamer2016loss, higgins2017darla, nair2018visual}.
~\cite{yarats2019improving}.
% , lange2010deep, shelhamer2016loss, higgins2017darla, nair2018visual}.
Another popular approach is contrastive representation learning in which there is an auxiliary task of selecting a matching representation (obtained for another transformation of the same input) among a number of alternatives \cite{laskin_srinivas2020curl}.
%which encourages the development of consistent representations for different augmentations of the same image input \cite{laskin_srinivas2020curl}.
%Contrastive learning is another recently popular unsupervised representation learning method that has found success in improving state representation learning for continuous control \cite{laskin_srinivas2020curl}.
More recently, image augmentations such as small random shifts have been found to enable data-efficient RL from images without any additional auxiliary loss functions~\cite{laskin2020reinforcement, yarats2021image}.

In this work, we represent the state of an actor-critic RL agent using a small set of ``feature points" extracted from high-dimensional image inputs. Feature points can be seen as spatial locations of features in the input image which are useful for the task at hand. Feature points could represent the locations of objects or object parts, or spatial relations between objects. For example, a feature point located between two object locations could track their relative motion. The motivation for using this type of representation is the intuition that the information in images consists of the appearance and geometry of the objects in the scene. Auxiliary tasks like autoencoding force the encoder to represent both types of information. However, what matters in many control problems is the geometry of the objects in the scene, not their exact appearance. Therefore, extracting the geometry information while neglecting the appearance information might lead towards better state representations.

% Motivate learning keypoints directly for RL  and list contributions
We use a simple feature point bottleneck on top of a convolutional encoder and train it end-to-end with an actor-critic algorithm, without any pre-training or additional supervision. %loss functions. % or access to the full state of the environment. 
Our approach, which we term as FPAC (feature point actor-critic), is robust to a different number of objects and object dynamics. While a convolutional encoder trained with the Soft Actor-Critic (SAC) algorithm \cite{haarnoja2018softa, haarnoja2018softb} is highly data-inefficient, the simple addition of a feature point bottleneck in FPAC greatly improves the learning performance, to attain data-efficiency and asymptotic performance competitive to state-of-the-art methods. The code to reproduce our FPAC experiments is available at \url{https://github.com/rinuboney/FPAC}.
% \textit{[link hidden for anonymity]}.
% provided in the supplementary material.
% 

%===============================================================================

\section{Related Work}
\label{sec:related}

% \textbf{Keypoint learning for continuous control from images}. 
Our work is closely related to prior works \cite{levine2016end,finn2016deep,cabi2019scaling} which learn feature points for continuous control from images. \cite{levine2016end, finn2016deep} train a non-linear neural policy with an intermediate feature point representation using supervised learning to imitate a local linear-Gaussian controller.
% which has access to the ground truth low-dimensional state of the environment. 
We use a similar architecture of the neural policy with a differential feature point bottleneck.
% as in \cite{levine2016end}.
%and we use the differential keypoint bottleneck proposed in \cite{finn2016deep}. \cite{levine2016end, finn2016deep} takes a guided policy search approach and models the environment as a time-varying linear dynamical system, learns local linear-Gaussian controllers, and trains a non-linear neural policy using supervised learning to imitate the linear-Gaussian controller from different initial states. In contrast, we take an actor-critic approach to the RL problem and directly learn the non-linear policy network. 
%While \cite{levine2016end, finn2016deep} assumes access to a known, fully observable state space of the environment, we assume access only to image observations. \cite{levine2016end, finn2016deep} pre-trains both the convolutional encoder (on a pose regression task) and the local linear-Gaussian controller and \cite{levine2016end} fine-tunes them in an end-to-end manner to learn task-relevant keypoints. In contrast, we show that it is possible to learn task-relevant keypoints from scratch without any such pre-training.
While \cite{levine2016end} assume access to the ground truth low-dimensional state of the environment, we
% In this work, we do not assume access to the environment state at training time and 
directly learn the control policy from pixels by feeding the extracted feature point representations to an actor-critic algorithm.
\cite{cabi2019scaling} learn feature points for batch reinforcement learning but rely on human demonstrations and manual reward annotation. We perform a study of feature point learning on popular continuous control tasks to show that it improves reinforcement learning from images, without any additional supervision, even in tasks with sparse rewards.

% \textbf{Unsupervised keypoint learning}. 
% Our work is related to recent works on unsupervised keypoint learning \cite{jakab2018unsupervised, kulkarni2019unsupervised}.
The feature point bottleneck mechanism proposed in \cite{levine2016end} has been used for unsupervised representation learning in several works. The works of \cite{finn2016deep, jakab2018unsupervised, zhang2018unsupervised} use it an auto-encoding framework to learn feature points in an unsupervised manner based on image generation.
% We use this keypoint extraction mechanism to extract keypoint locations but directly learn keypoints that are relevant for control.
\cite{kulkarni2019unsupervised} improved upon this by introducing a feature-transport mechanism to learn feature points that are more spatially aligned. Recently, \cite{gopalakrishnan2020unsupervised} proposed to learn object feature points based on local spatial predictability of image regions. Both \cite{kulkarni2019unsupervised} and \cite{gopalakrishnan2020unsupervised} demonstrate that feature points extracted using a pre-trained encoder serve as an effective state representation in some Atari games. We consider continuous control tasks and observe that pre-trained feature points fail to generalize well in some tasks while feature points learned end-to-end with RL perform better (see Fig.~\ref{fig:key}).
\cite{minderer2019unsupervised} learn stochastic feature point dynamics models in an unsupervised manner and demonstrate their efficacy on reward prediction in some continuous control tasks.
\cite{manuelli2020keypoints} trained dense image descriptions using self-supervised correspondence to extract keypoints and learn keypoint dynamics models for model-based control in real-world robotic manipulation.
While these prior works use generic tasks like image generation \cite{jakab2018unsupervised, zhang2018unsupervised, kulkarni2019unsupervised, gopalakrishnan2020unsupervised} or equivariance constraints \cite{thewlis2017unsupervised, zhang2018unsupervised, thewlis2019unsupervised} to learn feature points in an unsupervised manner, we aim to directly learn feature points that are relevant for control. We do not use any additional auxiliary losses that are specific to feature point learning but instead rely on end-to-end learning with RL losses
% to utilize the differentiable keypoint extraction mechanism
to learn feature points that are well-aligned for the control task at hand.

% keypoints in object detection?

% \textbf{Continuous control from images}. 
% There exist several state representation learning methods for continuous control from images. 
Any unsupervised visual representation learning method could be used to potentially improve RL from images. Compressing high-dimensional image observations using a pre-trained autoencoder to assist RL was first proposed in \cite{lange2010deep} and later improved in other works \cite{yarats2019improving, shelhamer2016loss, higgins2017darla, nair2018visual}.
Stable RL from images without any additional loss functions was demonstrated using the DDPG algorithm in \cite{tassa2020dmcontrol} using only the critic loss to update the convolutional encoder. \cite{yarats2019improving} jointly trained a regularized convolutional autoencoder with the SAC algorithm. \cite{laskin_srinivas2020curl} trained the convolutional encoder jointly using the SAC algorithm and an unsupervised contrastive loss function. 
Recently, \cite{laskin2020reinforcement} and \cite{yarats2021image} concurrently discovered that some image augmentations such as small random shifts enable data-efficient RL on image-based RL benchmarks, without any additional auxiliary loss functions. 
Learned latent dynamics models have also been successfully used for planning \cite{hafner2019learning} or to assist policy search \cite{hafner2019dream, ha2018recurrent} for continuous control from images.
These methods have been also successfully used to demonstrate data-efficient RL 
% Data-efficient RL from images has also been recently demonstrated 
in some real-world robot tasks \cite{singh2019, zhu2019ingredients, kendall2019learning, viitala2020learning}. In this paper, we aim to
improve sample-efficiency of RL by learning feature points %representations 
that distill the geometric information from images.

%===============================================================================

\section{Reinforcement Learning}

% Explain MDP
We formulate continuous control from images as a Markov decision process (MDP). An MDP consists of
a set of states $\mathcal{S}$,
a set of actions $\mathcal{A}$,
a transition probability function $\s_{t+1} \sim p(\cdot|\s_t, \act_t)$ that represents the probability of transitioning to a state $\s_{t+1}$ by taking action $\act_t$ in state $\s_t$ at timestep $t$,
a reward function $r_t = R(\s_t, \act_t)$ that provides a scalar reward for taking action $\act_t$ in state $\s_t$, and
a discount factor $\gamma \in [0, 1]$ to weigh future rewards.

The policy function $\act_t \sim \pi(\cdot|\s_t)$ of an RL agent defines the behavior of the agent: it is a mapping from the state to actions. 
% The state value function $V_\pi(s)$ is defined as the expected cumulative rewards by following policy $\pi$ from state $s$: $V^\pi(s) = \E [ \sum_{t=0}^\infty \gamma^t R(s_t, a_t) | s_0 = s, s_{t+1} \sim p(\cdot|s_t, a_t), a_t \sim \pi(s_t)]$. 
The goal in RL is to learn an policy function $\pi$ that maximizes the expected cumulative reward given by
\[
%\argmax_\theta % V_{\pi_\theta}(s)$ for every $s \in \mathcal{S}$.
\E_{\s_{t+1} \sim p(\cdot|\s_t, \act_t), \: \act_t \sim \pi(\s_t)} \left[ \sum_{t=0}^\infty \gamma^t R(\s_t, \act_t) %\mid \s_{t+1} \sim p(\cdot|\s_t, \act_t), \act_t \sim \pi(\s_t)
\right]
, \forall \s_0 \in \mathcal{S}.
\]

% Explain SAC and critic loss
In this paper, we solve the RL problem using the state-of-the-art Soft Actor-Critic (SAC) algorithm \cite{haarnoja2018softb}. SAC contains an actor network which produces stochastic policy $\pi$ and a critic network which implements the state-action value function $Q^\pi(\mathbf{s}, \mathbf{a})$, which is the expected cumulative reward after taking action $\mathbf{a}$ in state $\mathbf{s}$ and following $\pi$ thereafter. The policy $\pi$ is tuned to maximize an entropy-regularized RL objective
$
\E \left[ \sum_{t=0}^\infty \gamma^t (R(\s_t, \act_t) + \alpha\mathcal{H}(\pi(\cdot|\s_t)) \right]
$,
where
%$\theta$ denotes all learnable parameters of the agent and
$\alpha > 0$ is a learnable temperature parameter.
%to balance the joint optimization of the cumulative rewards and the entropy.
%
The critic network $Q$ is
%used to estimate the $Q^\pi(s, a)$ function of policy  $\pi$: the expected cumulative rewards after taking action $a$ in state $s$ and following $\pi$ thereafter. The critic network is
trained to satisfy the Bellman equation: $Q^\pi(\s_t, \act_t) = r_t + \gamma Q^\pi(\s_{t+1}, \pi(\s_{t+1}))$. It is updated by sampling transitions $(\s_t, \act_t, r_t, \s_{t+1})$ from a replay buffer $\mathcal{D}$ and minimizing the critic loss
\begin{equation*}
J_Q = \mathbb{E}_{(\s_t, \act_t, r_t, \s_{t+1}) \sim \mathcal{D}} \left[ (Q(\s_t, \act_t) - r_t - \gamma V(\s_{t+1}))^2 \right] \,,    
%\label{eq:critic}
\end{equation*}
where the soft value function $V(\s_t) = \mathbb{E}_{\act_t \sim \pi(\cdot|\s_t)} \left[ Q(\s_t, \act_t) - \alpha \log\pi(\act_t|\s_t) \right]$
%(that is, the expected cumulative rewards obtained by following policy $\pi$ from state $\s$)
is approximated using a Monte Carlo estimate.
%In this paper, we use this critic loss to learn keypoint representations that are relevant for RL from images.
We refer the reader to \cite{haarnoja2018softb} for more information about SAC.

%===============================================================================

\section{Learning Feature Point State Representation}

\begin{figure*}[t]
\centering
\includegraphics[width=0.9\textwidth, trim=50 280 100 61, clip]{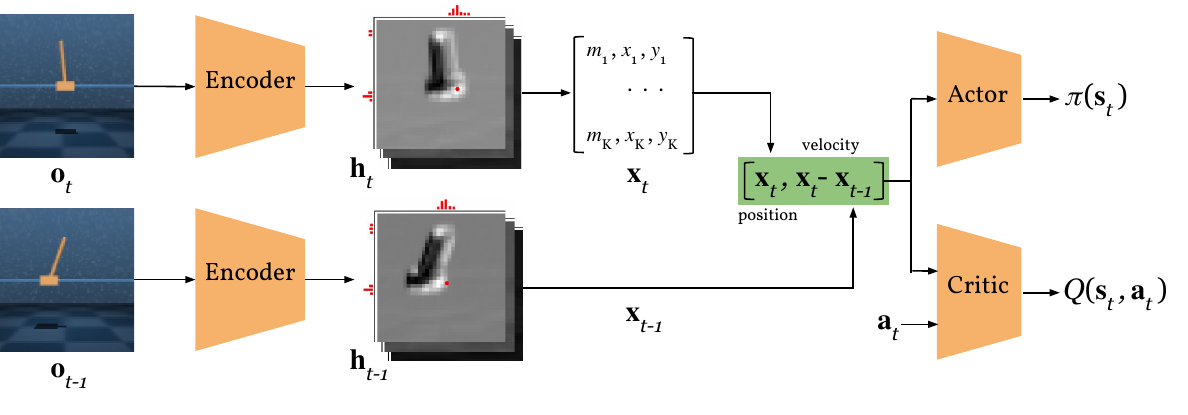}
\caption{
\textbf{FPAC Architecture}. At time-step $t$, a convolutional encoder is used to extract $K$ feature maps $\h_t$ from image observations $\obs_t$. The features maps are considered as probability distributions of feature point locations and we extract $K$ feature point locations and features $\x_t = [(m_1, x_1, y_1),\ldots,(m_K, x_K, y_K)]$ from these $K$ probability distributions $\h_t$ (Equations \eqref{eq:key2d} and \eqref{eq:keyp}). We similarly compute $\h_{t-1}$ and $\x_{t-1}$ from $\obs_{t-1}$ and represent the current state as a concatenation of feature point position $\x_t$ and feature point velocity $\x_t - \x_{t-1}$. This feature point state representation is given to the actor and critic networks. The feature points (that is, the weights of the convolutional encoder) are learned based on the gradients from the critic loss. %\eqref{eq:critic}.
}
\label{fig:arch}
\end{figure*}

In this paper, we assume that the state $\s_t$ is a stack of two consecutive images, that is $\s_t = [\obs_t, \obs_{t-1}]$.
We represent the state $\s_t$ as a collection of feature points extracted from input images $\obs_t$, $\obs_{t-1}$ with a differentiable feature point extractor. The extracted feature points are then used as the inputs of the actor and the critic heads of the SAC agent (see Fig.~\ref{fig:arch}).
%In this paper, we explore the possibility of representing the state $\mathbf{s}_t$ of the agent using a set of keypoints automatically extracted from input images. Our proposed architecture consists of a differentiable keypoint extractor that produces state $\mathbf{s}_t$ which is used as the input of the SAC agent (see Fig.~\ref{fig:arch}).

We define a feature point as a triplet $(x_k, y_k, m_k)$ where $x_k$ and $y_k$ are the 2D coordinates of the feature point and $m_k \in (-1, 1)$ is a scalar feature which can, for example, encode the presence of the feature point in an image.
Each input image $\obs$ is processed into a set of $K$ feature points
with a differentiable feature point bottleneck $\Phi$:
%$\x = \Phi(\obs)$ where $\Phi()$ is implemented as a differentiable keypoint extractor and $\x = [(m_1, x_1, y_1),\ldots,(m_K, x_K, y_K)]$.
\[
  \x = \Phi(\obs) = [(x_1, y_1, m_1),\ldots,(x_K, y_K, m_K)].
\]
%
%Since the relative positions of keypoints might generalize better than absolute positions, we use relative keypoint positions $\bar{\x} = [(m_1, x_1-\bar{x}, y_1-\bar{y}),\ldots,(m_K, x_K-\bar{x}, y_K-\bar{y})]$, where $\bar{x}$ and $\bar{y}$ are the mean $x$ and $y$ coordinates of all $K$ keypoints.
% We found experimentally that relative positions of keypoints
%\[\bar{\x} = [(x_1-\bar{x}, y_1-\bar{y},m_1),\ldots,(x_K-\bar{x}, y_K-\bar{y},m_K)],\]
% where $\bar{x}$ and $\bar{y}$ are the mean $x$ and $y$ coordinates of all $K$ keypoints, generalize better than absolute positions.
%For continuous control from images, we define the state as a stack of consecutive image observations, that is $\s_t = [\obs_t, \obs_{t-1}, \obs_{t-2}, \ldots]$, where $\obs \in \mathbb{R}^{H' \times W' \times C}$.
%
%Therefore, 
We use the vector
$
  [\x_t, \x_t - \x_{t-1}]
$ 
as the inputs of the actor and the critic heads,
where $\x_t = \Phi(\obs_t)$, $\x_{t-1} = \Phi(\obs_{t-1})$ and $\x_t - \x_{t-1}$ encodes the feature point velocities. While we do not introduce explicit constraints to encourage temporal consistency of keypoints, this feature point velocity term could implicitly encourage that. 
% (See Fig.~\ref{fig:velocity} in the Appendix for an ablation study of this velocity term).
%In this paper, we assume that the states consist of stacks of two consecutive images, that is, $\s_t = [\obs_t, \obs_{t-1}]$.

% Motivate
% How to extract and represent the right geometric information for RL from images?
%In this paper, we explore the possibility of directly learning keypoint representations that are relevant for continuous control from images. Keypoints are a flexible and structured representation of the geometry of the objects in the image.
% Describe setting
%We are interested in learning a function $\x = \Phi(\obs)$ from high-dimensional image observations $\obs$ to a list of $K$ keypoint locations $\x = [(m_1, x_1, y_1),\ldots,(m_K, x_K, y_K)]$, where $x_k$ and $y_k$ are the 2D coordinates of keypoint $k$ and $m_k$ is a scalar keypoint feature (which could for example represent if keypoint $k$ is absent in the image observation $\obs$).
% that could represent if keypoint $k$ is visible in image $\obs$. 

% Explain keypoint representation bottleneck layer
We implement function $\x = \Phi(\obs)$
using the differentiable feature point bottleneck proposed by \cite{levine2016end}.
We use a convolutional network $f$ to process image $\obs\in \mathbb{R}^{H' \times W' \times C}$ into $K$ feature maps of shape $H \times W$: $\h = f(\obs) \in \mathbb{R}^{H \times W \times K}$. Let % $\h^k(x, y)$ 
$h(x, y, k)$ denote the value of the $k$-th channel of $\h$ at pixel $(x, y)$. 
% Let $\sigma$ be the softmax function $\sigma(\mathbf{z}) = e^\mathbf{z} / \sum_{z\in\mathbf{z}} e^z$. 
The feature point coordinates $(x_k, y_k)$ can be taken as the expected values of the pixel coordinates:
\begin{equation}
% (x^k, y^k) = \sum_{u=1}^H \sum_{v=1}^W (u, v) \sigma(\h^k)(u, v) \,,
\begin{bmatrix}
x_k \\y_k
\end{bmatrix} = \sum_{x=1}^W \sum_{y=1}^H \begin{bmatrix}
x\\y
\end{bmatrix} p_k(x, y) \,,
\label{eq:key2d}
\end{equation}
where the expectation is computed using distribution $p_k(x, y)$ produced using the softmax function:
\[
p_k(x,y) = \frac{
\exp(h(x,y, k))
}{
\sum_{x',y'} \exp(h(x',y',k))
}.
\]
The scalar feature $m_k$ is computed as the $\tanh$-activated mean value of the feature maps:
\begin{equation}
m_k = \tanh\Big( \frac{1}{H W} \sum_{x=1}^W \sum_{y=1}^H h(x, y, k) \Big) \,.
\label{eq:keyp}
\end{equation}

In practice, we compute the coordinates $(x_k, y_k)$ using a separable variant of \eqref{eq:key2d} that is more efficient and also performed well in our experiments.
% to compute the coordinates $(x^k, y^k)$ more efficiently.
% : we perform mean pooling over one dimension of the 2D feature maps and then compute softmax over the other dimension. That is:
% \begin{equation}
% % x^k = \sum_{v=1}^W v \sigma\Big( \frac{1}{\beta H} \sum_{u=1}^H \h^k(u, v) \Big)(v) 
% % \text{ and }
% % y^k = \sum_{u=1}^H u \sigma\Big( \frac{1}{\beta W} \sum_{v=1}^W \h^k(u, v) \Big)(u)  \,,
% p_k(x,y) = \begin{bmatrix}
% \frac{
% \exp(\mean(\h, y)(k, x) / \beta)
% }{
% \sum_{x'} \exp(\mean(\h, y)(k, x') / \beta)
% }
% &
% \frac{
% \exp(\mean(\h, x)(k, y) / \beta)
% }{
% \sum_{y'} \exp(\mean(\h, x)(k, y') / \beta)
% }
% \end{bmatrix} \,,
% \label{eq:key1d}
% \end{equation}
% where $\mean(\h, d)$ applies mean pooling on $\h$ over coordinate $d$ and $\beta$ is a softmax temperature parameter (we use $\beta=0.5$ in all our experiments). 
%
% The keypoint presence feature $m_k$ could represent if keypoint $k$ is absent in the image observation $\obs$ and we model this
%We model the keypoint feature $m_k$ as the $\tanh$-activated mean value of the feature maps:
%\begin{equation}
%m_k = \tanh\Big( \frac{1}{H W} \sum_{x=1}^W \sum_{y=1}^H h(x, y, k) \Big) \,.
%\label{eq:keyp}
%\end{equation}
% The agent could use this keypoint presence feature $m_k$ to represent the presence or absence of keypoint $k$ in the image observation. 
% While keypoints could be absent when the objects in the scene moves out of camera view (like in Cartpole Swingup task considered in this paper), we observe that use of this feature only has a minor impact on performance.
For example, the $x$ coordinate is computed as
\begin{equation}
x_k = \sum_{x=1}^W x p_k(x),
\qquad
p_k(x) = \frac{
\exp(\frac{1}{\beta}g(k,x))
}{
\sum_{x'} \exp(\frac{1}{\beta} g(k,x'))
},
\label{eq:key1d}
\end{equation}
where $g(k,x)$ is the mean-pooled value of $h(x,y,k)$ along dimension $y$:
\begin{align*}
g(k,x) = \frac{1}{H} \sum_{y=1}^H h(x,y,k).
\end{align*}
The $y$ coordinate is computed similarly.
We use $\beta=0.5$ in all our experiments.

%The convolutional network $f$ and Equations \ref{eq:key1d} and \ref{eq:keyp} together compose the function $\x = \Phi(\obs)$ that maps image observations $\obs$ to keypoint locations and features $\x$.
% Describe full state: keypoint positions and velocity
%At time-step $t$, the RL agent receives the current state $\s_t$ of the environment. In this paper, we assume that the states consist of stacks of two consecutive images, that is, $\s_t = [\obs_t, \obs_{t-1}]$. We use $\Phi$ to extract $K$ keypoint locations and features $\x_t$ and $\x_{t-1}$ from $\obs_t$ and $\obs_{t-1}$ respectively. Since the relative positions of keypoints might generalize better than absolute positions, we compute relative keypoint positions $\bar{\x} = [(m_1, x_1-\bar{x}, y_1-\bar{y}),\ldots,(m_K, x_K-\bar{x}, y_K-\bar{y})]$, where $\bar{x}$ and $\bar{y}$ are the mean $x$ and $y$ coordinates of all $K$ keypoints. We additionally compute the keypoint velocities $\x_t - \x_{t-1}$. The concatenation of all values in relative keypoint positions $\bar{\x}_t$ and keypoint velocities $\x_t - \x_{t-1}$ forms the keypoint representation used in this paper and we provide this concatenated vector to the actor and critic networks. 

The extracted feature points depend only on the weights of the convolutional encoder $f$. These weights are updated using the gradients of the critic loss 
% \eqref{eq:critic}
so that the agent directly learns the feature point locations that are relevant for the RL task. We call our agent FPAC (feature point actor-critic) and its complete architecture is illustrated in Fig.~\ref{fig:arch}.
% The PyTorch code for extracting feature points from convolutional feature maps is provided in Listing~\ref{lst:FPAC} in the Appendix.

%===============================================================================

\section{Experimental Results}
\label{sec:result}

\begin{figure*}[t!]
\centering
\includegraphics[width=0.9\textwidth, trim=5 10 8 5, clip]{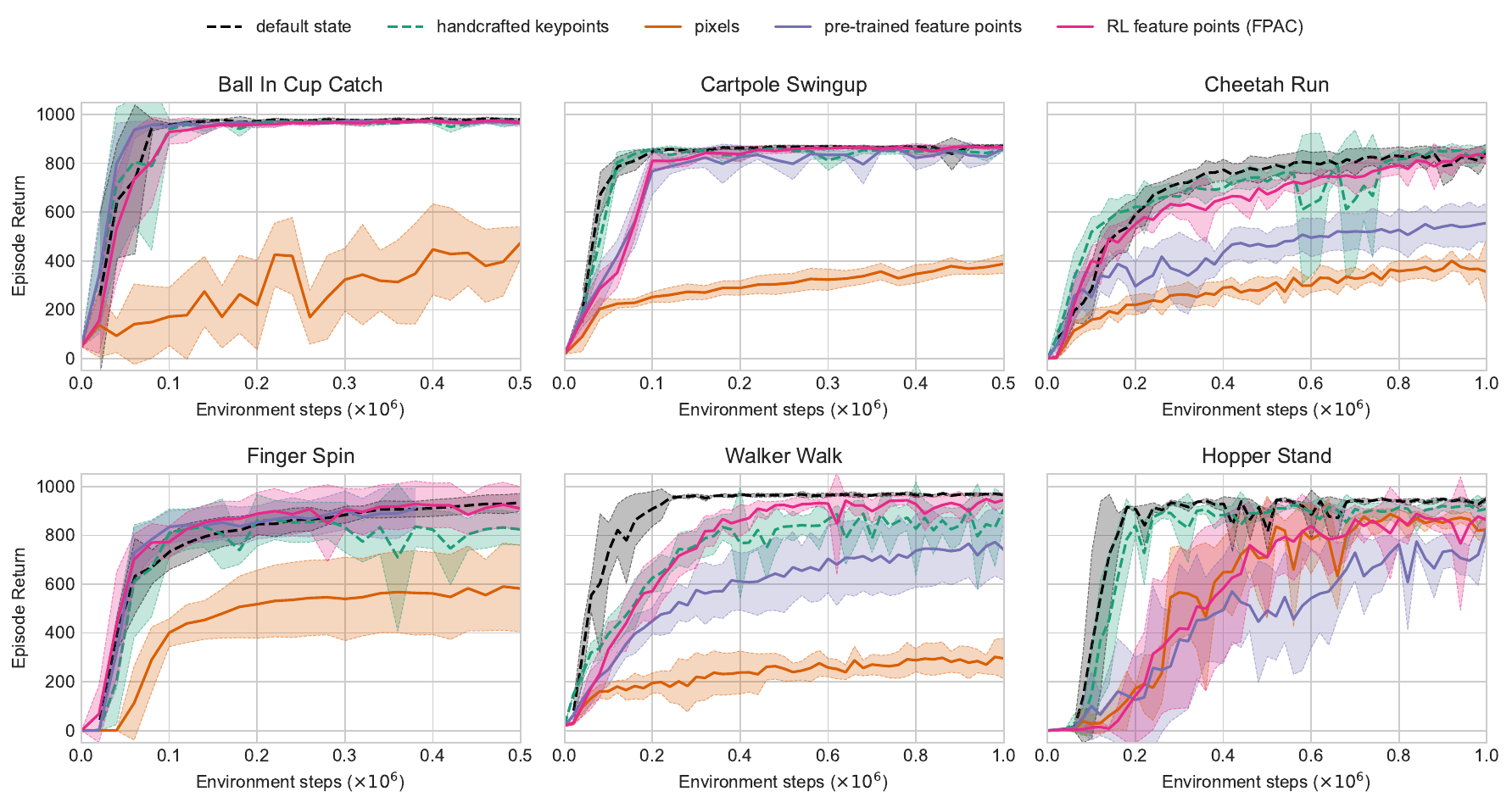}
\caption{
% FPAC works.
Evaluation of the efficacy of spatial coordinate representations on the PlaNet benchmark using the SAC algorithm. 
%Dashed lines represent agents that learn from low-dimensional state representations, while solid lines represent agents that learn from image observations. 
%
SAC with \emph{handcrafted keypoints} (extracted from the simulator) performs almost as well as SAC with fine-tuned \emph{default state} representations defined in DeepMind Control Suite, which suggests that 2D keypoints can be effective for state representation for control.
%We show that 2D keypoints are an effective state representation as the SAC algorithm can learn all tasks from \emph{handcrafted keypoints} (extracted from the simulator), with performance comparable to that of learning from fine-tuned \emph{default state} representations defined in DeepMind Control Suite.
%
While SAC from \emph{pixels} is unable to learn, %directly from \emph{pixels},
adding a feature point extraction bottleneck at the output of the convolutional encoder (\emph{FPAC}) significantly improves performance. Using \emph{pre-trained feature points} representations (pre-trained on extra data of 10k steps) performs well in simple tasks like Cartpole Swingup but does not perform as well in more challenging tasks like Cheetah Run, Walker Walk, and Hopper Stand.
We plot the mean and standard deviation of 6 runs of all agents.
}
\label{fig:key}
\end{figure*}

\begin{figure}[t!]
\centering
\includegraphics[width=\linewidth, trim=5 10 8 5, clip]{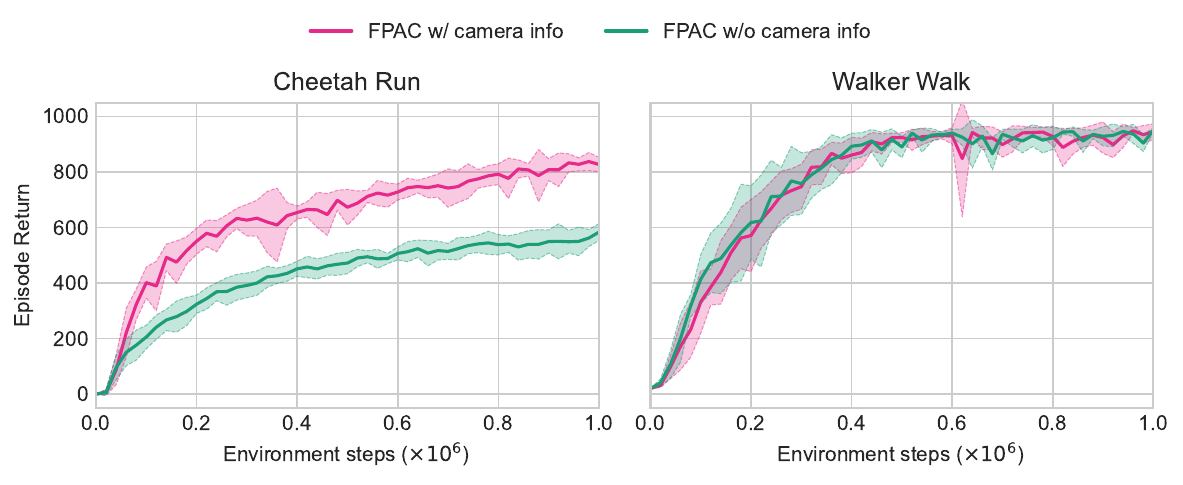}
\caption{Ablation study on the use of camera movement information. We observe that the camera movement information significantly helps in Cheetah Run while it doesn't impact the performance in Walker Walk. We hypothesize that this is due to the fact that the ``cheetah" always stays in the center of the image and the only motion cue in these observations is the lightly colored checkboard pattern on the floor.}
\label{fig:camera-info}
\end{figure}

\begin{figure*}[t!]
\centering
\includegraphics[width=0.9\textwidth, trim=5 10 8 5, clip]{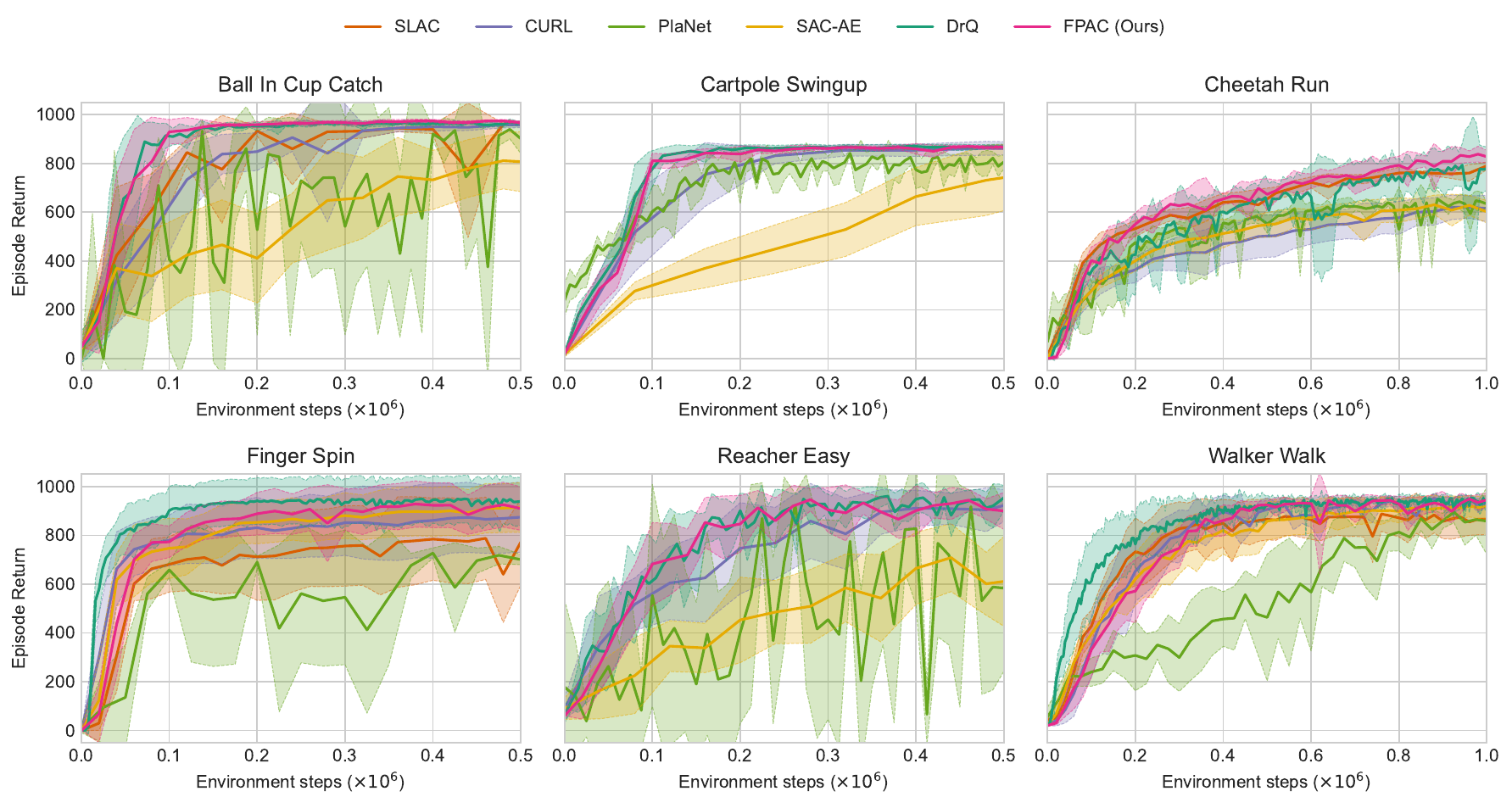}
\caption{
Comparison of FPAC to prior methods on the PlaNet benchmark. FPAC performs competitively to the state-of-the-art DrQ, CURL and SLAC methods.
% and better than all the other methods on most tasks. 
We plot the mean and standard deviation of 10 runs of all agents. Results of other methods taken from \cite{yarats2021image}.
}
\label{fig:comparison}
\end{figure*}

\begin{figure*}[t!]
\centering
\includegraphics[width=1\textwidth, trim=5 10 8 5, clip]{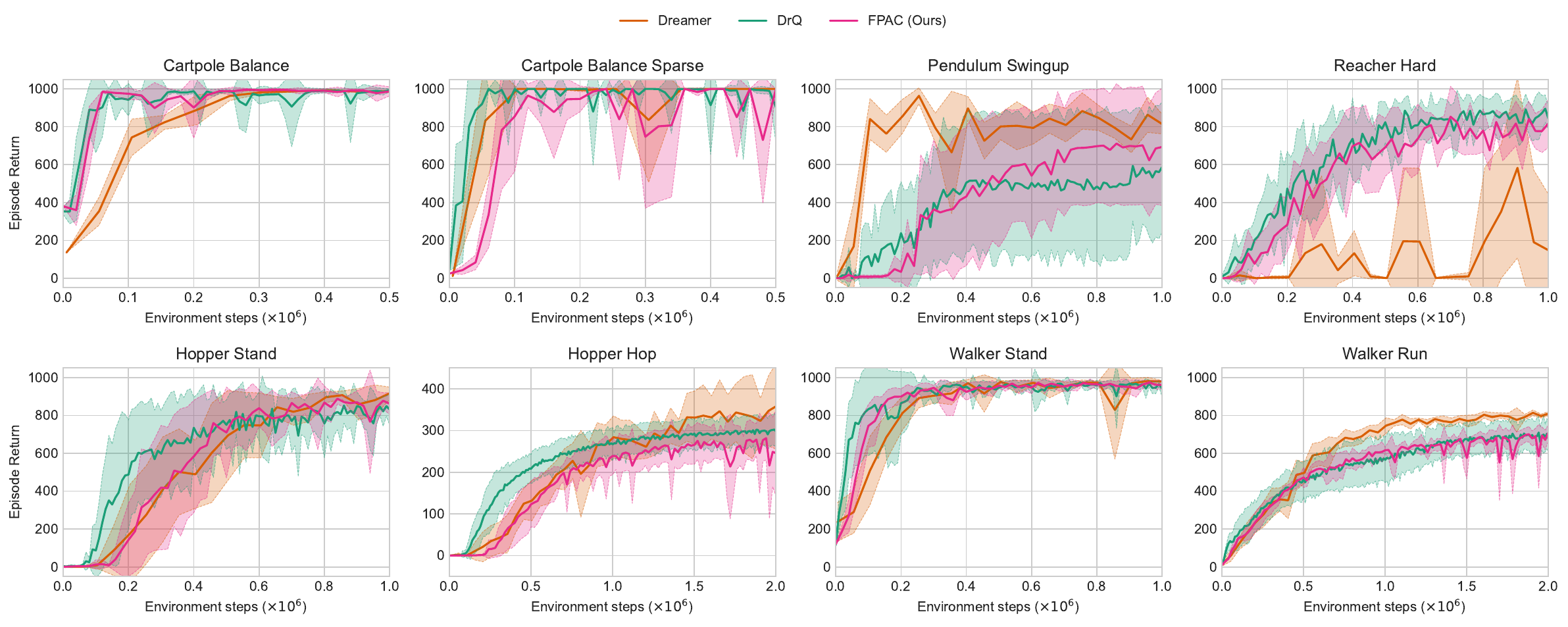}
\caption{
Comparison of FPAC to prior methods on eight additional tasks from the Dreamer benchmark. FPAC performs competitively to the state-of-the-art DrQ and Dreamer methods. We plot the mean and standard deviation of 10 runs of all agents. Results of other methods taken from \cite{yarats2021image}.
}
\label{fig:dreamer}
\end{figure*}

\begin{figure}[t!]

\begin{subfigure}{.325\linewidth}
\caption{Ball in Cup}
\centering\vspace{-2mm}
\includegraphics[width=1.0\linewidth, trim=110 45 110 45, clip]{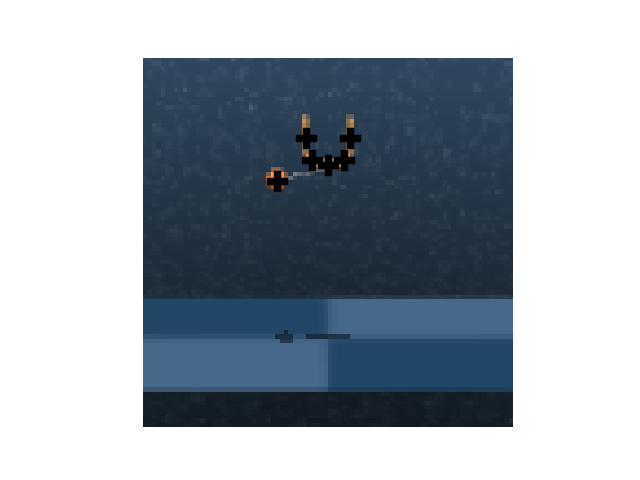}
\end{subfigure}
\begin{subfigure}{.325\linewidth}
\centering
\caption{Cartpole}\vspace{-2mm}
\includegraphics[width=1.0\linewidth, trim=110 45 110 45, clip]{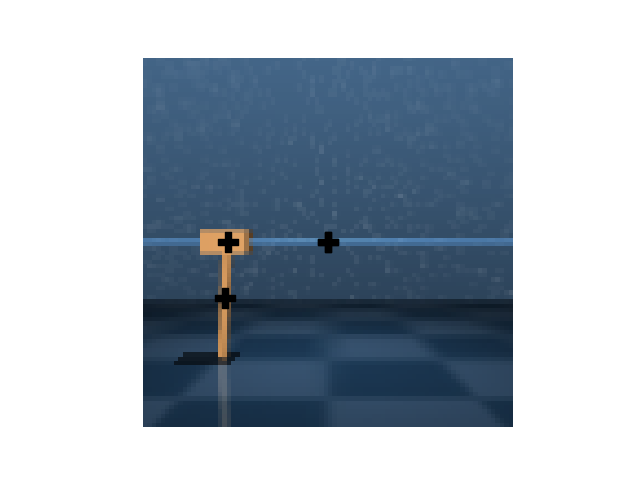}
\end{subfigure}
\begin{subfigure}{.325\linewidth}
\centering
\caption{Finger}\vspace{-2mm}
\includegraphics[width=1.0\linewidth, trim=110 45 110 45, clip]{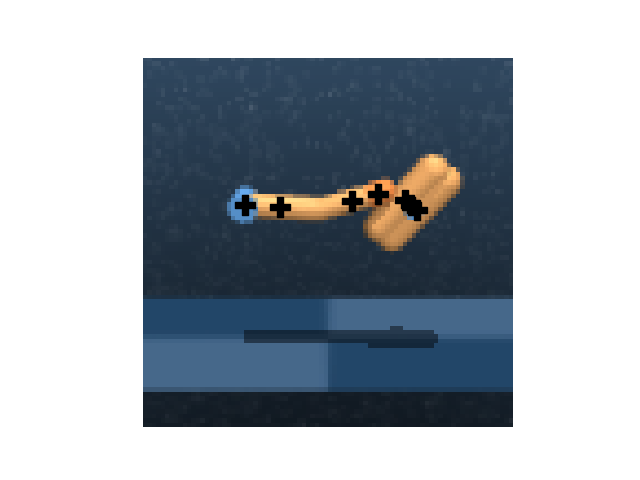}
\end{subfigure}

\vspace{1pt}

\begin{subfigure}{.325\linewidth}
\centering
\includegraphics[width=1.0\linewidth, trim=110 45 110 45, clip]{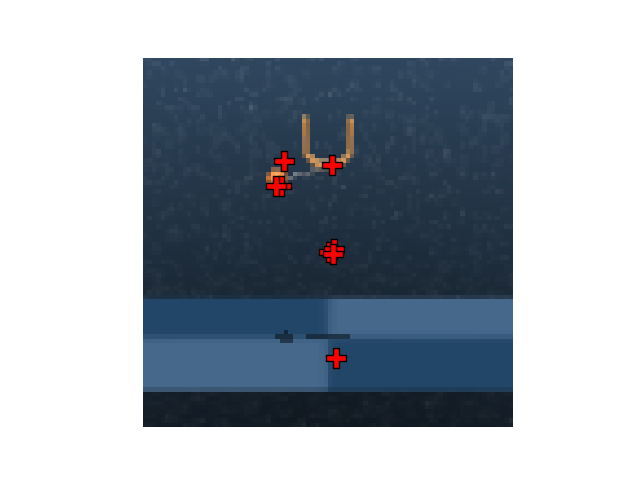}
\end{subfigure}
\begin{subfigure}{.325\linewidth}
\centering
\includegraphics[width=1.0\linewidth, trim=110 45 110 45, clip]{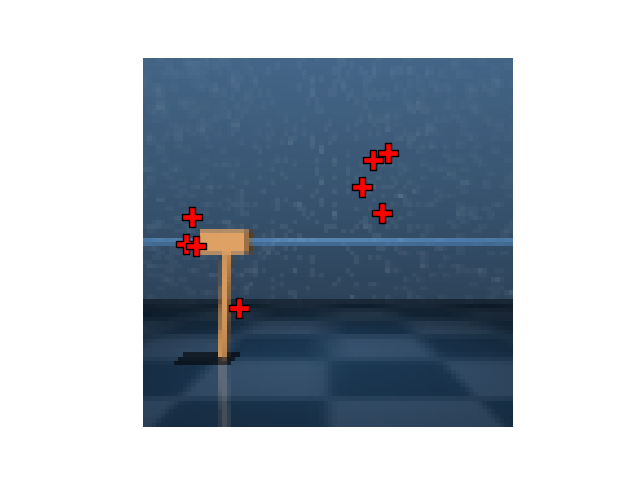}
\end{subfigure}
\begin{subfigure}{.325\linewidth}
\centering
\includegraphics[width=1.0\linewidth, trim=110 45 110 45, clip]{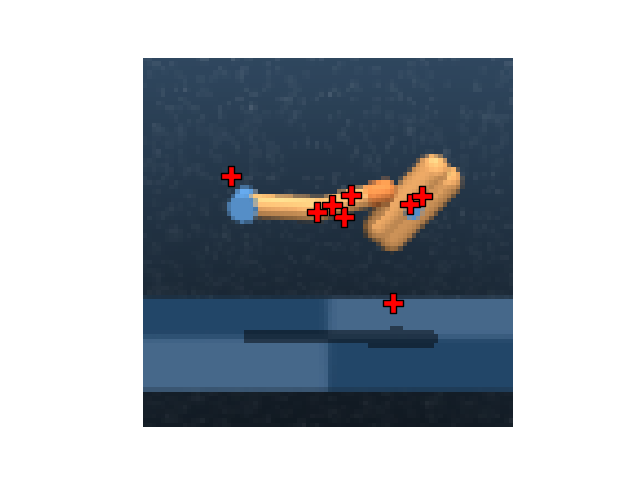}
\end{subfigure}

\vspace{1pt}

\begin{subfigure}{.325\linewidth}
\centering
\includegraphics[width=1.0\linewidth, trim=125 35 93 55, clip]{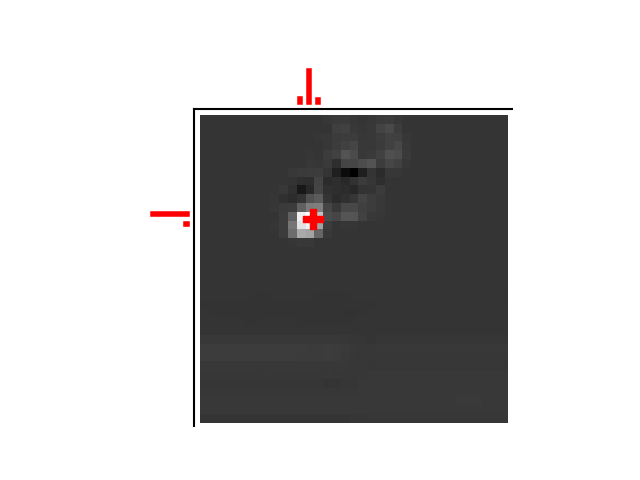}
\end{subfigure}
\begin{subfigure}{.325\linewidth}
\centering
\includegraphics[width=1.0\linewidth, trim=125 35 93 55, clip]{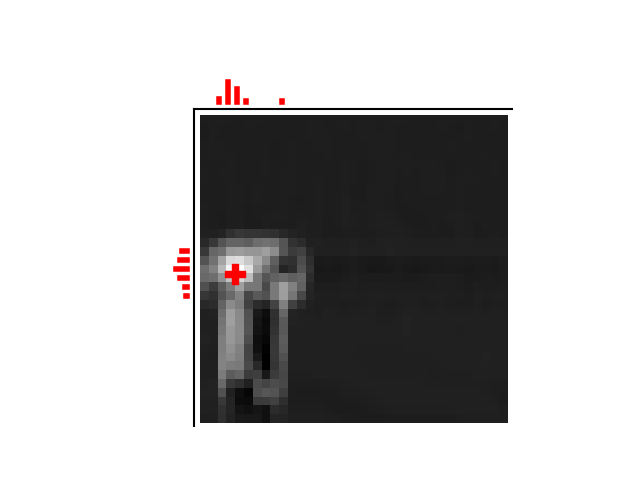}
\end{subfigure}
\begin{subfigure}{.325\linewidth}
\centering
\includegraphics[width=1.0\linewidth, trim=125 35 93 55, clip]{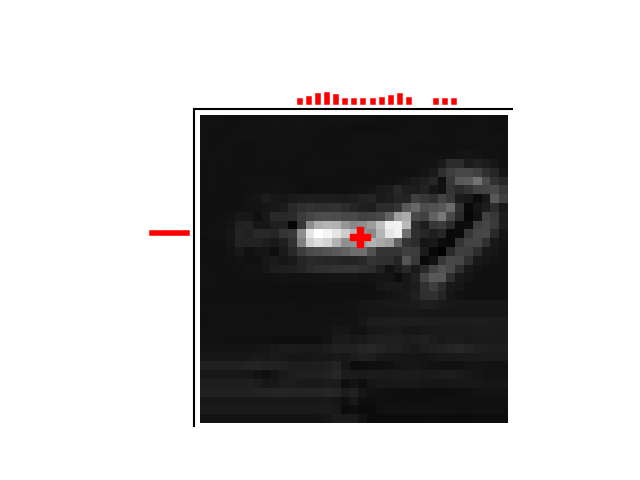}
\end{subfigure}

\vspace{1pt}

\begin{subfigure}{.325\linewidth}
\centering
\includegraphics[width=1.0\linewidth, trim=125 35 93 55, clip]{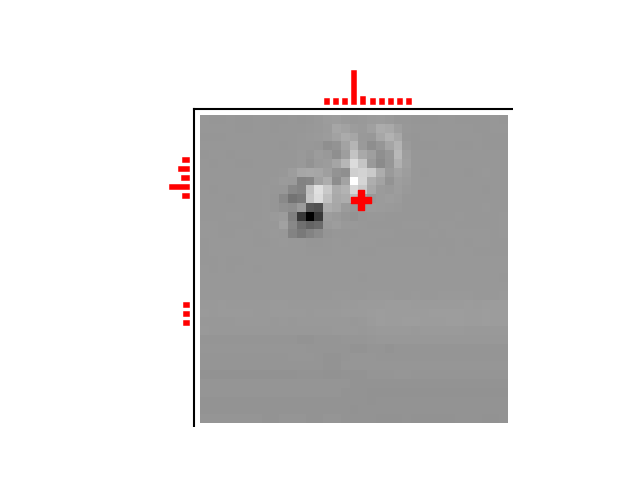}
\end{subfigure}
\begin{subfigure}{.325\linewidth}
\centering
\includegraphics[width=1.0\linewidth, trim=125 35 93 55, clip]{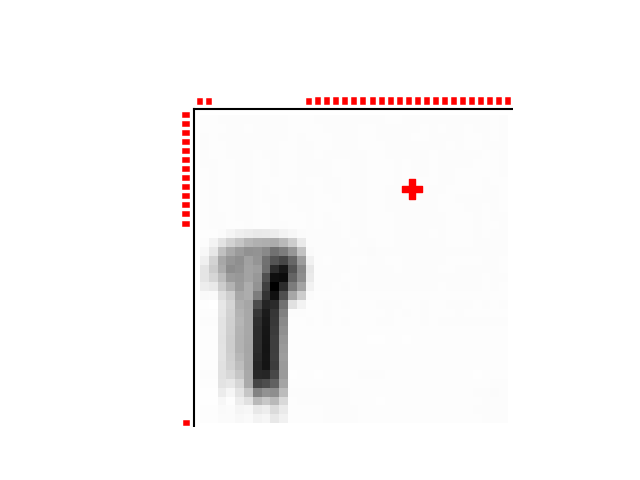}
\end{subfigure}
\begin{subfigure}{.325\linewidth}
\centering
\includegraphics[width=1.0\linewidth, trim=125 35 93 55, clip]{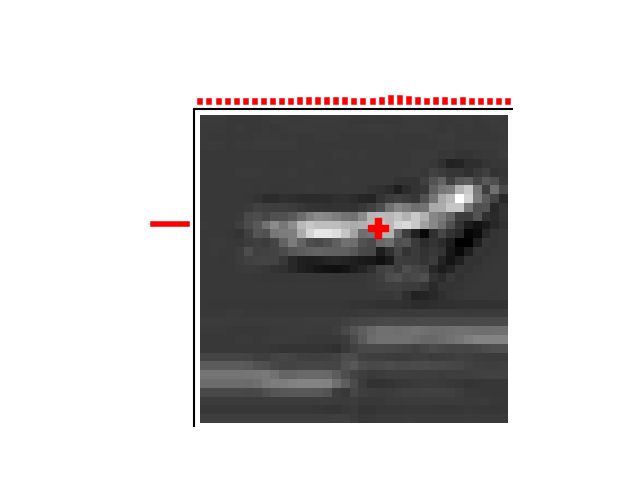}
\end{subfigure}

\caption{
\emph{Row 1}: Example frames from three tasks from the PlaNet benchmark. The black crosses represent the handcrafted keypoints extracted from the simulator.
\emph{Row 2}: Feature points learned by FPAC with $K=8$.
\emph{Rows 3 and 4}: Some feature maps $h(\cdot,\cdot,k)$ produced by the convolutional FPAC encoder along with the probability distributions of the feature point location (Equation~\ref{eq:key1d}). Brighter pixels indicate larger values.
FPAC can learn to represent parts of an object (row 3), multiple objects (by activating the locations of multiple objects as, for example, in row 4 for Finger and Ball in Cup) or the background (row 4 for Cartpole).
}
\label{fig:keypoints}

\end{figure}

\begin{figure*}[tp]

\begin{subfigure}{.162\textwidth}
\caption{$t$}\vspace{-2mm}
\centering
\includegraphics[width=1.0\linewidth, trim=125 35 93 55, clip]{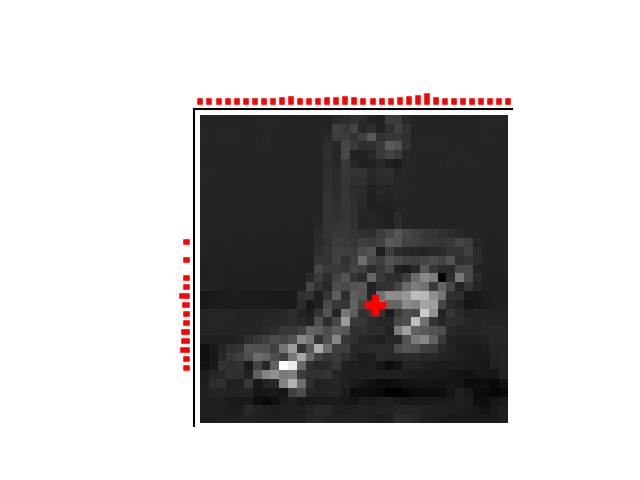}
\end{subfigure}
\begin{subfigure}{.162\textwidth}
\centering
\caption{$t+5$}\vspace{-2mm}
\includegraphics[width=1.0\linewidth, trim=125 35 93 55, clip]{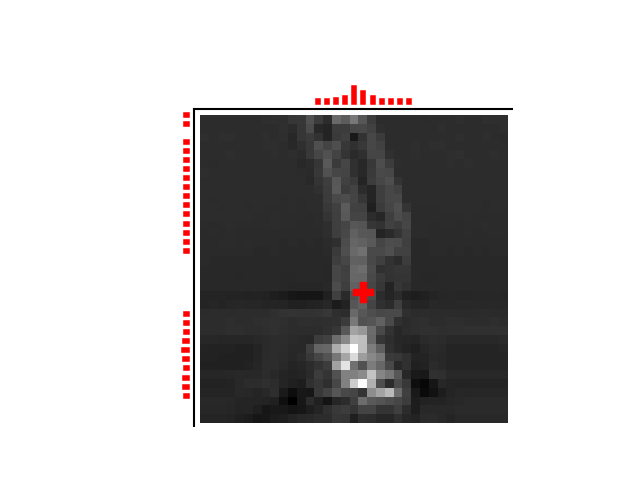}
\end{subfigure}
\begin{subfigure}{.162\textwidth}
\centering
\caption{$t+10$}\vspace{-2mm}
\includegraphics[width=1.0\linewidth, trim=125 35 93 55, clip]{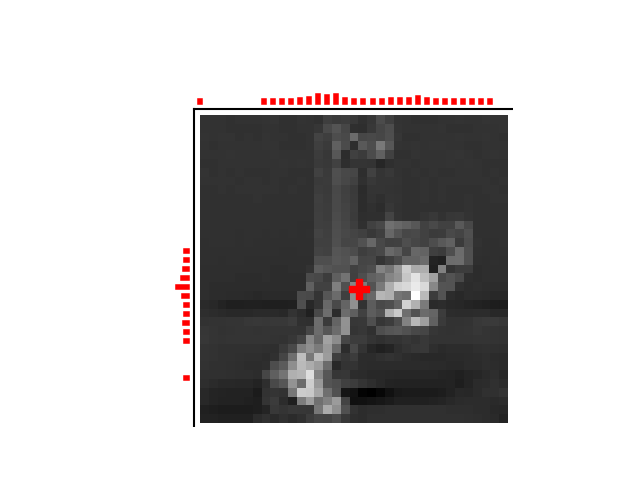}
\end{subfigure}
\begin{subfigure}{.162\textwidth}
\centering
\caption{$t+15$}\vspace{-2mm}
\includegraphics[width=1.0\linewidth, trim=125 35 93 55, clip]{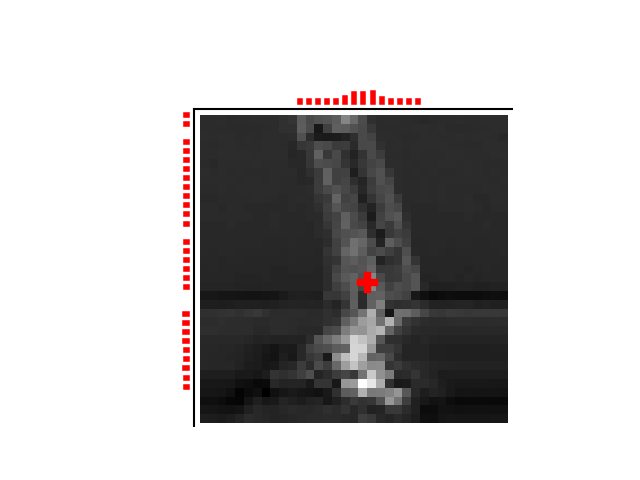}
\end{subfigure}
\begin{subfigure}{.162\textwidth}
\centering
\caption{$t+20$}\vspace{-2mm}
\includegraphics[width=1.0\linewidth, trim=125 35 93 55, clip]{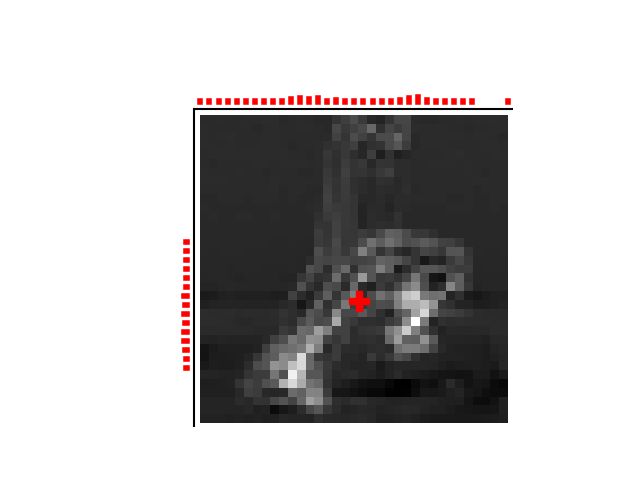}
\end{subfigure}
\begin{subfigure}{.162\textwidth}
\centering
\caption{$t+25$}\vspace{-2mm}
\includegraphics[width=1.0\linewidth, trim=125 35 93 55, clip]{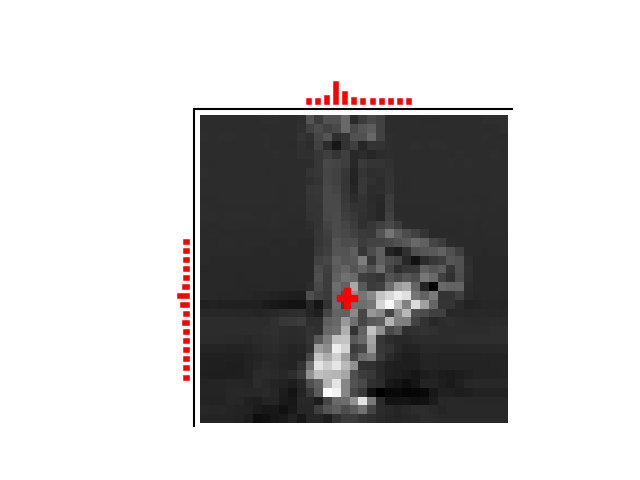}
\end{subfigure}

\vspace{1pt}

\begin{subfigure}{.162\textwidth}
\includegraphics[width=1.0\linewidth, trim=125 35 93 55, clip]{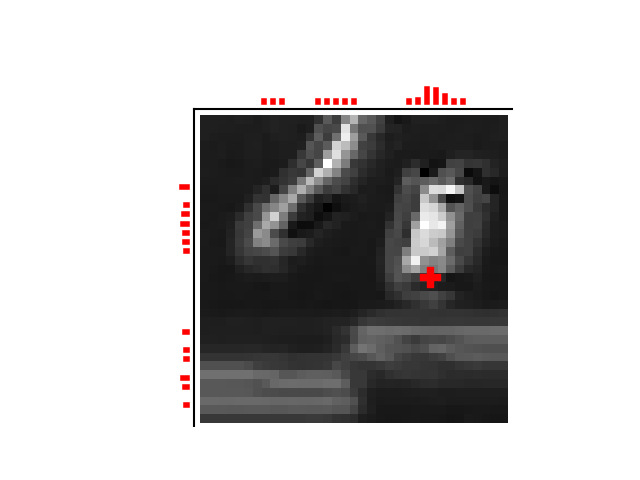}
\end{subfigure}
\begin{subfigure}{.162\textwidth}
\includegraphics[width=1.0\linewidth, trim=125 35 93 55, clip]{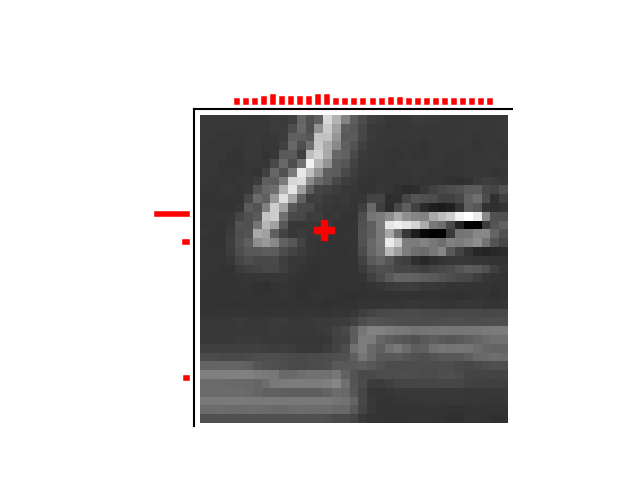}
\end{subfigure}
\begin{subfigure}{.162\textwidth}
\includegraphics[width=1.0\linewidth, trim=125 35 93 55, clip]{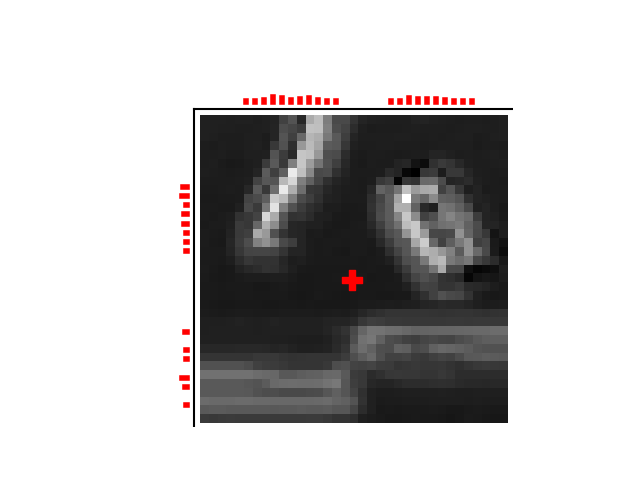}
\end{subfigure}
\begin{subfigure}{.162\textwidth}
\includegraphics[width=1.0\linewidth, trim=125 35 93 55, clip]{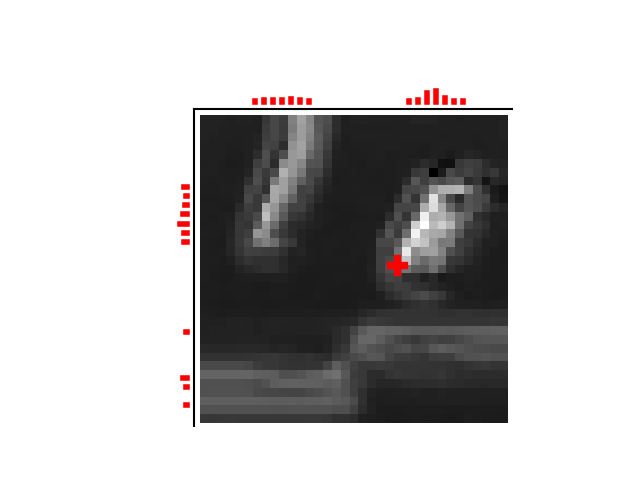}
\end{subfigure}
\begin{subfigure}{.162\textwidth}
\includegraphics[width=1.0\linewidth, trim=125 35 93 55, clip]{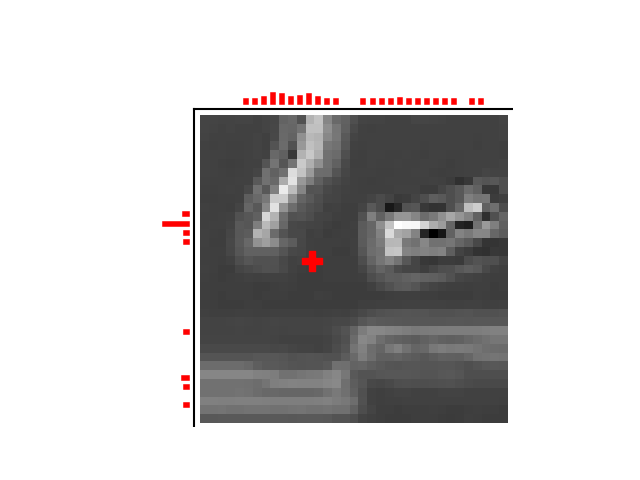}
\end{subfigure}
\begin{subfigure}{.162\textwidth}
\includegraphics[width=1.0\linewidth, trim=125 35 93 55, clip]{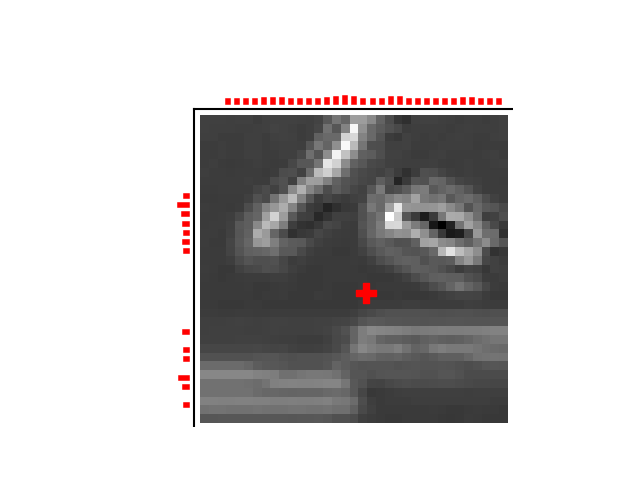}
\end{subfigure}

% \vspace{1pt}

% \begin{subfigure}{.162\textwidth}
% \includegraphics[width=1.0\linewidth, trim=125 35 93 55, clip]{corl_analysis/cheetah_run/190_0.png}
% \end{subfigure}
% \begin{subfigure}{.162\textwidth}
% \includegraphics[width=1.0\linewidth, trim=125 35 93 55, clip]{corl_analysis/cheetah_run/200_0.png}
% \end{subfigure}
% \begin{subfigure}{.162\textwidth}
% \includegraphics[width=1.0\linewidth, trim=125 35 93 55, clip]{corl_analysis/cheetah_run/210_0.png}
% \end{subfigure}
% \begin{subfigure}{.162\textwidth}
% \includegraphics[width=1.0\linewidth, trim=125 35 93 55, clip]{corl_analysis/cheetah_run/220_0.png}
% \end{subfigure}
% \begin{subfigure}{.162\textwidth}
% \includegraphics[width=1.0\linewidth, trim=125 35 93 55, clip]{corl_analysis/cheetah_run/230_0.png}
% \end{subfigure}
% \begin{subfigure}{.162\textwidth}
% \includegraphics[width=1.0\linewidth, trim=125 35 93 55, clip]{corl_analysis/cheetah_run/240_0.png}
% \end{subfigure}

\caption{
Dynamics of feature maps of keypoints that learn to represent multiple objects. Brighter pixels indicate larger values.
\emph{Row 1}: a keypoint that tracks the front and back legs in the Walker Walk task.
\emph{Row 2}: a keypoint that tracks the finger and rotatable body, relative to the floor, in the Finger Spin task.
% \emph{Row 3}: a keypoint that tracks the relative motion of the front and beck ends of the torso in the Cheetah Run task.
}
\label{fig:keypoint-dynamics}

\end{figure*}

\begin{figure*}[t!]
\centering
\includegraphics[width=1\textwidth, trim=8 5 12 5, clip]{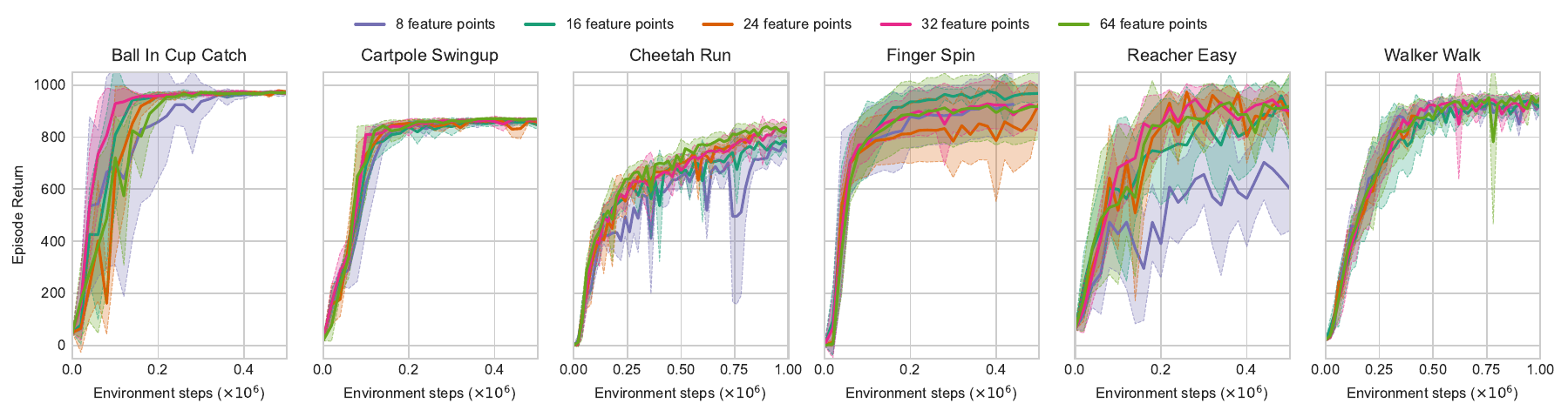}
\caption{
FPAC learns robustly with a different number $K$ of extracted feature points. FPAC achieves similar asymptotic performance on all tasks, except Reacher Easy.
%, with only 8 keypoints and performs similarly with 16 or 32 keypoints.
We plot the mean and standard deviation of 6 runs of all agents.
}
\label{fig:num_key}
\end{figure*}

% Describe PlaNet benchmark
In this section, we evaluate our FPAC method on a set of image-based continuous control tasks from the DeepMind Control Suite (DMC)~\cite{tassa2020dmcontrol}.
We first evaluate it on six tasks from the PlaNet benchmark \cite{hafner2019learning}
% The six tasks from the considered benchmark are 
% (illustrated in the first row of Fig.~\ref{fig:keypoints}) 
and further on eight additional tasks from the Dreamer benchmark \cite{hafner2019dream}.
%, along with example image observations and handcrafted keypoints (row 1).
The tasks present qualitatively different learning challenges like the robot moving out of camera view (Cartpole), sparse rewards  (Reacher, Ball in Cup), contacts between objects (Finger, Cheetah, Hopper, Walker), and a large number of joints (Cheetah, Walker).
Following all prior works on the PlaNet benchmark \cite{yarats2019improving, laskin_srinivas2020curl, yarats2021image, hafner2019learning, lee2020stochastic}, we use different action repeat values for each task. % (listed in Table~\ref{tab:hyperparams})
%, as originally used by the PlaNet method \cite{hafner2019learning} and all other subsequent methods evaluated on the benchmark.
and report the true environment steps (which is invariant to the action repeat parameter) in all our experiments. The Dreamer benchmark uses a constant action repeat of 2 steps on all tasks.

In the first experiment, we test whether spatial coordinate representations such as feature points can be an effective form of state representation for continuous control.
Fig.~\ref{fig:key} shows the learning curves of the different versions of the SAC agent:
\begin{itemize}
\item Agent that learns from raw \emph{pixels}. This is a SAC agent that learns low-dimensional state representations from a stack of image observations with a convolutional encoder based on the critic loss \cite{haarnoja2018softb}. The only difference between this agent and our FPAC agent is the additional feature point bottleneck used in FPAC.

\item Agent that uses the low-dimensional \emph{default state} from DMC. This includes information like robot pose, joint positions, and joint velocities, and was fine-tuned separately for each task by \cite{tassa2020dmcontrol}.

\item Agent that uses ground-truth locations of \emph{handcrafted keypoints} that we manually extract from the simulator. We compute the keypoints by taking the 3D locations of the center of all objects in the environment and projecting them to the 2D pixel space of the default camera defined for all tasks. 
We found experimentally that relative positions of keypoints $\bar{\x} = [(x_1-\bar{x}, y_1-\bar{y},m_1),\ldots,(x_K-\bar{x}, y_K-\bar{y},m_K)],$ where $\bar{x}$ and $\bar{y}$ are the mean $x$ and $y$ coordinates of all $K$ keypoints, generalize better than absolute positions with handcrafted keypoints and use this in our experiments.
The first row of Fig.~\ref{fig:keypoints} shows examples of the handcrafted keypoints for all tasks in the PlaNet benchmark. 
% The Python code for extracting such pre-defined keypoints from the DeepMind Control Suite is provided in Listing~\ref{lst:gt_key} in the Appendix.

\item Our \emph{FPAC} agent which learns feature point representations of images from scratch using a convolutional feature point extractor updated using the critic loss.

\item Agent that uses \emph{pre-trained feature points}. We train a self-supervised feature points encoder from 10k images collected using a random policy. Pre-training is done similarly to \cite{jakab2018unsupervised} by minimizing the reconstruction loss. The pre-trained feature points are used as the SAC inputs.
\end{itemize}

% For agents with handcrafted keypoints, we found experimentally that relative positions of keypoints $\bar{\x} = [(x_1-\bar{x}, y_1-\bar{y},m_1),\ldots,(x_K-\bar{x}, y_K-\bar{y},m_K)],$ where $\bar{x}$ and $\bar{y}$ are the mean $x$ and $y$ coordinates of all $K$ keypoints, generalize better than absolute positions and we use this in all experiments with handcrafted keypoints. We found our FPAC agent to work well with both absolute and relative keypoints (See Section~\ref{sec:relative} in the Appendix for an ablation study) and use absolute keypoints for pre-trained keypoints and FPAC experiments as it works better in all tasks.

% We explain this in detail in the Appendix and also perform an ablation study of this use of extra camera information in Fig.~\ref{fig:camera-info}.

One can see that the SAC algorithm can learn effectively from the handcrafted keypoints, with similar data-efficiency and asymptotic performance to that of learning from the default state, except for Walker Walk where it performs slightly worse. Note that the default state space is fine-tuned for each task such that RL algorithms can successfully learn from them \cite{tassa2020dmcontrol} and the center locations of objects might not be the optimal spatial coordinate representation for continuous control in all tasks.
Our FPAC agent achieves similar data-efficiency and asymptotic performance as SAC from the handcrafted keypoints. FPAC performs worse in Cartpole and Hopper but better in Walker and Finger.
SAC from pixels performs poorly in all tasks except Hopper. FPAC performs significantly better by simply introducing an extra feature points bottleneck layer.

We observe that while SAC from pre-trained feature points performs better than SAC from pixels, it performs worse than FPAC in Cheetah and Walker. As the RL agent learns, it actively visits new states and the pre-trained feature points fail to generalize to these new states (see Section~\ref{sec:generalization}).
FPAC directly learns the feature points relevant for the RL task and performs as well as SAC from the handcrafted keypoints and almost as well as SAC from the default state. These results suggest that end-to-end learning of feature points without any additional pre-training works at least as well as using a pre-trained feature point extractor on all considered tasks.

In Cheetah, Walker and Hopper tasks, for agents that use handcrafted keypoints, pre-trained feature points, and RL feature points (FPAC), we use additional information about the movements of the camera to translate the feature points in the past frame to the same coordinates as the current frame. 
This allows the agent to separate the movements of the robot and the camera and only use what is relevant for the task (that is, only the robot movement). Note that this is also possible in real-world applications by using computer vision approaches to track the movement of the camera. Also, we use this information only in the agents that use feature point representations because inserting this information to other agents is not trivial. We perform an ablation study of this use of extra camera information in Fig.~\ref{fig:camera-info}.

In Fig.~\ref{fig:comparison} and Fig.~\ref{fig:dreamer}, we compare our FPAC method to prior methods on tasks from the PlaNet and Dreamer benchmarks:
%To demonstrate the effectiveness of our FPAC method, we now compare the learning performance of FPAC against several prior methods:
%The comparison methods are:
\begin{itemize}
\item 
PlaNet \cite{hafner2019learning} and Dreamer \cite{hafner2019dream} are model-based RL algorithms that learn latent dynamics models of the environment and computes actions by planning with them.
% the learned model. % using cross-entropy method.

\item SLAC \cite{lee2020stochastic} also learns a latent dynamics model of the environment but learns policy and value functions on top of the latent representation, using the SAC algorithm.
\item SAC-AE \cite{yarats2019improving} learns a regularized convolutional autoencoder jointly with the SAC algorithm. % using reconstruction and critic losses.
\item CURL \cite{laskin_srinivas2020curl} learns a convolutional encoder using an unsupervised constrastive loss jointly with the SAC algorithm.
\item DrQ \cite{yarats2021image} averages the $Q$ predictions and targets over different random shifts of image observations to train a convolutional encoder, based only on the critic loss.
\end{itemize}

% Describe comparison results
% Following prior works, we compare the data-efficiency and asymptotic performance of FPAC based on returns on 100k and 500k steps of the PlaNet benchmark in Table~\ref{tab:result}. FPAC performs competitively to the state-of-the-art methods.
% FPAC performs competitively to the state-of-the-art DrQ method and better than all the other methods on most tasks. FPAC outperforms all methods in the Ball in Cup Catch and Cheetah Run tasks. % FPAC can perform robustly across tasks with a different number of objects and different types of motion.
% To further test the robustness of FPAC, we compare it to prior methods on eight additional tasks from the Dreamer benchmark in Fig.~\ref{fig:dreamer}. 

FPAC performs well on all tasks, with sample-efficiency and asymptotic performance comparable to the state-of-the-art DrQ and Dreamer methods and better than all the other methods on most tasks.
The 14 different tasks we considered in our experiments have a different number of objects and object dynamics. FPAC can learn feature points to represent them, to perform robustly well on all tasks.

% Describe plot of learned keypoints
We plot the feature points learned by FPAC along with a few selected feature maps produced by the convolutional encoder in Fig.~\ref{fig:keypoints}. % In Ball in Cup, Cartpole, and Reacher, feature points mainly 
FPAC learns to represent the locations of the relevant objects in the scene. 
% FPAC also learns to use feature points to represent multiple objects. This is achieved by activating the locations of multiple objects in the feature map. The location of such feature points would lie between the multiple objects. % and allows the agent to track the distance between them, that is, their relative motion. 
% In Cheetah, Finger, and Walker, FPAC uses most feature points to represent multiple objects. This allows the agent to track how one object moves relative to another.
% In Cheetah, the same part on the front and back legs tend to be represented by the same keypoint so that it tracks how the part in one leg moves relative to the same part in the other leg. Similarly, in Finger, FPAC learns to represent the finger and the rotatable body and also the distance between them.
We observe that FPAC also learns feature points to represent multiple objects that are relevant to the control task (see Fig.~\ref{fig:keypoint-dynamics}).
% , for example, the two legs in Walker (see row 1 in Fig.~\ref{fig:keypoint-dynamics}), the finger and the rotatable body in Finger (see row 2 in Fig.~\ref{fig:keypoint-dynamics}), and parts of the front and back body in Cheetah. 
% (see row 3 in Fig.~\ref{fig:keypoint-dynamics}).
The feature point representations learned by neural networks do not typically correspond to explainable visual cues. Learning more human interpretable feature points is an important topic for future research.

% Describe robustness of num_keypoints
We analyze the robustness of FPAC to the number of learned feature points, which is a hyperparameter. The results of our experiments are shown in Fig.~\ref{fig:num_key}. We observe that FPAC is robust to the number of feature points and performs similarly well on all tasks with 8, 16, 24, 32, or 64 feature points, except Reacher Easy where FPAC performs worse with 8 feature points. So, given enough capacity, FPAC is robust and performs well across all tasks. 

All model-free algorithms considered in this paper (SAC from pixels, SAC-AE, CURL, DrQ, and FPAC) use the same reinforcement learning algorithm (SAC) and base network architectures: a four-layer convolutional encoder  (with 32 channels, kernel size 3, and a stride of 2 on the first layer) and shallow feedforward actor-critic networks. All hyperparameters used in our experiments are listed in Table~\ref{tab:hyperparams}.
% Following prior works, we also compare the data-efficiency and asymptotic performance based on episode returns on 100k and 500k steps of the PlaNet benchmark in Table~\ref{tab:result} in the Appendix.

\begin{table}[t]
\caption{
Hyperparameters used in our experiments. The hyperparameters are common across all tasks, except the initial learning rate for Reacher.
}
\label{tab:hyperparams}
\centering
\begin{tabular}{l|l}
\toprule
Hyperparameter & Value \\
\midrule
Observation size & (84, 84) \\
Replay buffer capacity & 100000 \\
Batch size & 128 \\
\multirow{2}{*}{Learning rate} & 1e-3, first 200k steps of Reacher \\
 & 3e-4, othwerise \\
Optimizer & Adam \\
Evaluation episodes & 10 \\
Discount factor $\gamma$ & 0.99 \\
Initial random steps & 1000 \\
Initial temperature &  0.1 \\
Target update rate $\tau$ & 0.01 \\
Target update frequency & 2 \\
Actor update frequency & 2 \\
Frame stack & 2 \\
MLP hidden layers & 2 \\
MLP hidden units & 1024 \\
Non-linearity & Swish \\
Number of feature points & 32 \\
Feature point temperature & 0.5 \\
\bottomrule
\end{tabular}
\end{table}

% We further analyze the design choices and robustness of our FPAC agent in detail in the Appendix. %~\ref{sec:ablation}.
% We observe that our FPAC method is robust to design and implementation choices such as the use of scalar feature $m_k$, use of feature point velocity term, use of relative spatial coordinates, and the use of separable spatial softmax implementation.

% Describe analysis of the robustness of learned keypoints

% We further analyze the robustness of the policy to the learned keypoints in Fig.~\ref{fig:drop} in the Appendix and observe that the performance of FPAC agents degrades smoothly as the number of dropped keypoints increases and the agent is still able to reasonably perform well even after dropping 25\% of the keypoints. We also perform an ablation study on the use of keypoint scalar feature $m_k$ in Fig.~\ref{fig:feature} in the Appendix.

Our FPAC method is easy to implement and fast to run. We measure an overall training time of 70 minutes 41 seconds to train a SAC from pixels agent and 76 minutes 56 seconds to train our FPAC agent for 500 episodes on the Cartpole Swingup task. We report an average of 10 runs on an NVIDIA V100 GPU. The only difference between the SAC from pixels agent and our FPAC agent is the additional feature point bottleneck used in FPAC. The additional feature point bottleneck makes our approach only negligibly slower (while performing significantly better) than SAC from pixels (which does not learn well).
We provide the PyTorch code for computing feature points from convolutional feature maps in Listing~\ref{lst:FPAC}.

\begin{listing}[tb]
\caption{PyTorch pseudocode for computing feature points from convolutional feature maps}
\label{lst:FPAC}
\begin{lstlisting}[language=Python]
def compute_feature_points(feature_maps, temp=1):
  """
  Compute feature points from feature_maps of shape [batch_size, num_points, H, W]
  """
  coords = torch.linspace(-1, 1, 
    steps=feature_maps.size(-1))[None, None, :]

  def compute_coord(other_axis):
    # Mean pooling
    logits_1d = feature_maps.mean(other_axis)
    # Compute feature point probabilities
    probs = F.softmax(logits_1d / temp, dim=-1)
    # Compute expectation of the distribution
    mean = (coords * probs).sum(-1)
    return mean

  # Compute feature point locations
  xs = compute_coord(-2)
  ys = compute_coord(-1)
  # Compute scalar feature
  ms = torch.flatten(feature_maps, -2, -1)
  ms = torch.tanh(ms).mean(-1)
  return xs, ys, ms
\end{lstlisting}
\end{listing}

% \begin{listing*}[tb]
% \caption{PyTorch code for computing feature points from convolutional feature maps}
% \label{lst:FPAC}
% \begin{lstlisting}[language=Python]
% def compute_feature_points(feature_maps, temperature=1):
%     """
%     Compute feature point locations and features from 
%     feature_maps of shape [batch_size, num_points, H, W]
%     """
%     device = feature_maps.device
%     maps_size = feature_maps.size(-1)
%     coords = torch.linspace(-1, 1, maps_size, device=device)

%     def compute_coord(other_axis):
%         # Mean pooling
%         logits_1d = feature_maps.mean(other_axis)
%         # Compute feature point probabilities
%         key_probs = F.softmax(logits_1d / temperature, dim=-1)
%         # Compute expectation of the distribution
%         mean = (coords[None, None, :] * key_probs).sum(-1)
%         return mean

%     # Compute feature point locations
%     xs = compute_coord(-2)
%     ys = compute_coord(-1)
%     # Compute scalar feature
%     ps = torch.tanh(torch.flatten(feature_maps, -2, -1).mean(-1))
%     return xs, ys, ps
% \end{lstlisting}
% \end{listing*}

%===============================================================================

\section{Ablation studies}
\label{sec:ablation}

\begin{figure}[t]
\centering
\includegraphics[width=\linewidth]{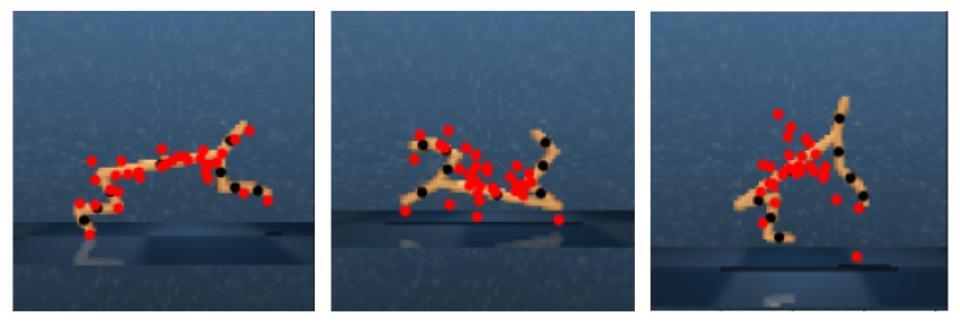}
\caption{Failure of pre-trained feature points on Cheetah Run. \emph{Left}: self-supervised training learns to represent all relevant parts of the cheetah robot. \emph{Middle and Right}: pre-trained feature point predictions fail to generalize to unseen observations. As an RL agent learns, it actively visits new states and it is crucial to use representations that generalize to new states or to continuously learn them.}
\label{fig:pretrain-failure}
\end{figure}

\begin{figure*}[t]
\centering
\includegraphics[width=0.45\textwidth, trim=0 0 0 60, clip]{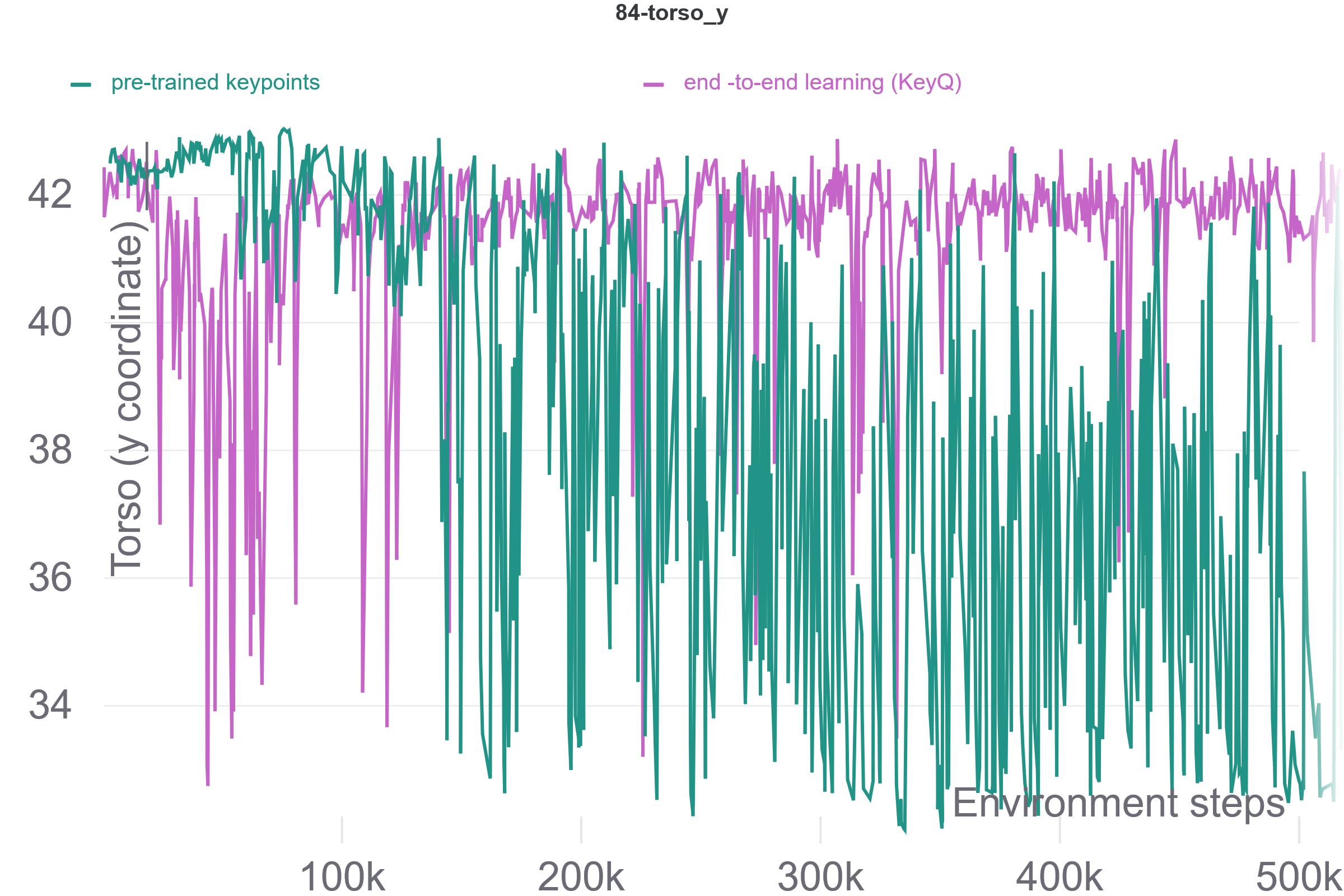}
\includegraphics[width=0.45\textwidth, trim=0 0 0 60, clip]{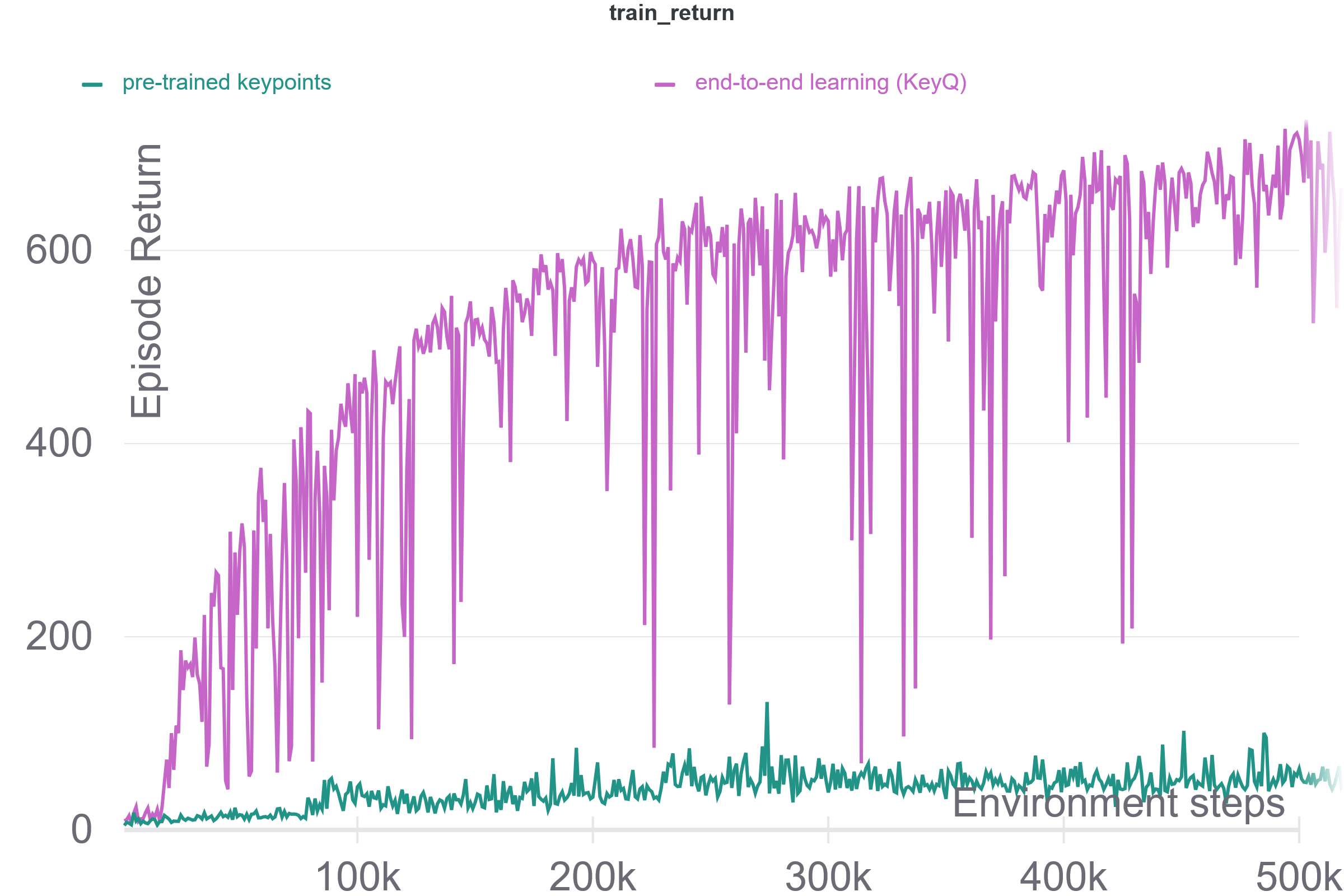}
\caption{Analysis on the failure of pre-trained feature points (learned using self-supervised training) on Cheetah Run. \emph{Left}: We plot the y-coordinate of the torso of the cheetah throughout two training runs on the Cheetah Run task. Note that we plot the position in pixels coordinates of $84 \times 84$ image observations. A high to low drop in this value indicates that the cheetah has flipped upside down. We can observe that the agent with pre-trained feature points continues to flip the cheetah and never recovers from it but the end-to-end learning agent quickly recovers from it and learns to solve the task. \emph{Right}: We plot the returns of the same agents. As evident from plot of the torso position, we can observe that the agent with pre-trained feature points fails to progress on the task.}
\label{fig:pretrain-ablation}
\end{figure*}

\begin{figure*}[t]
\centering
\includegraphics[width=1\textwidth, trim=5 5 5 5, clip]{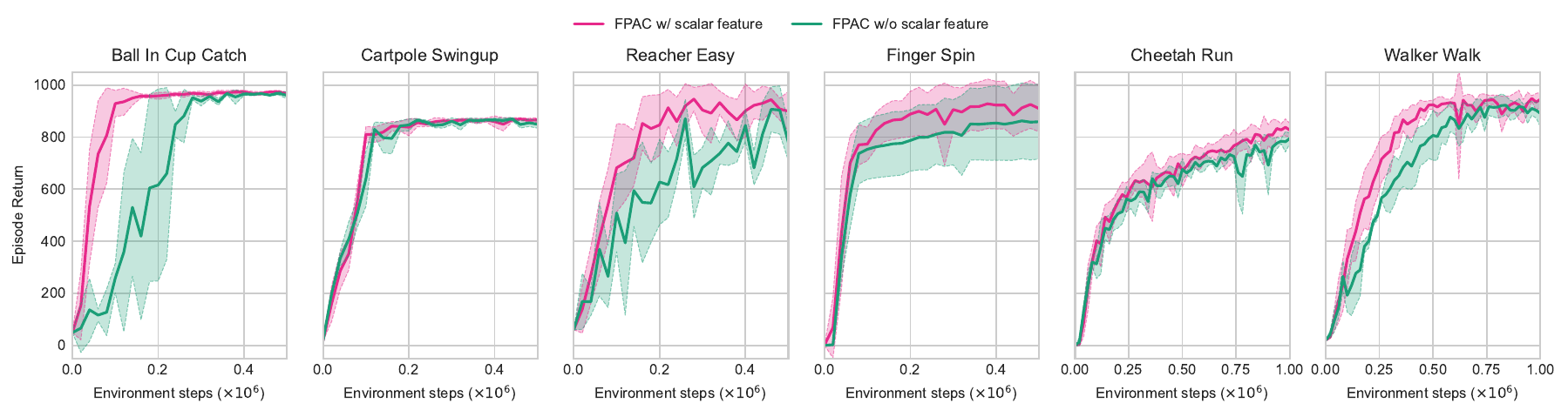}
\caption{
Impact of feature point scale feature $m_t$ on learning performance of FPAC.
}
\label{fig:feature}
\end{figure*}

\begin{figure*}[t]
\centering
\includegraphics[width=1\textwidth, trim=5 5 5 5, clip]{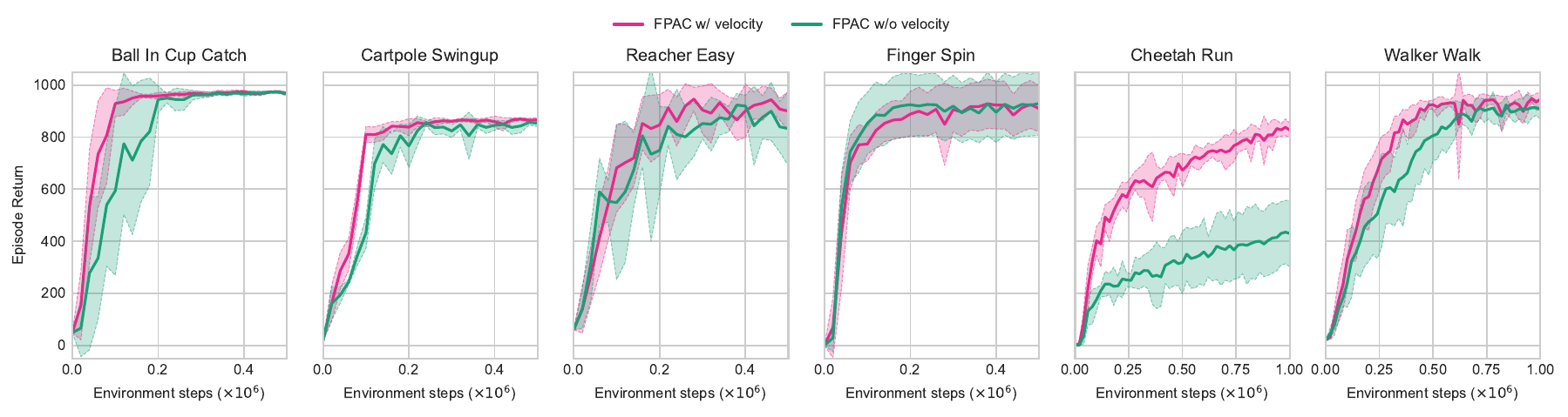}
\caption{
Impact of feature point velocity term on learning performance of FPAC.
}
\label{fig:velocity}
\end{figure*}

\begin{figure*}[t!]
\centering
\includegraphics[width=1\textwidth, trim=5 5 5 5, clip]{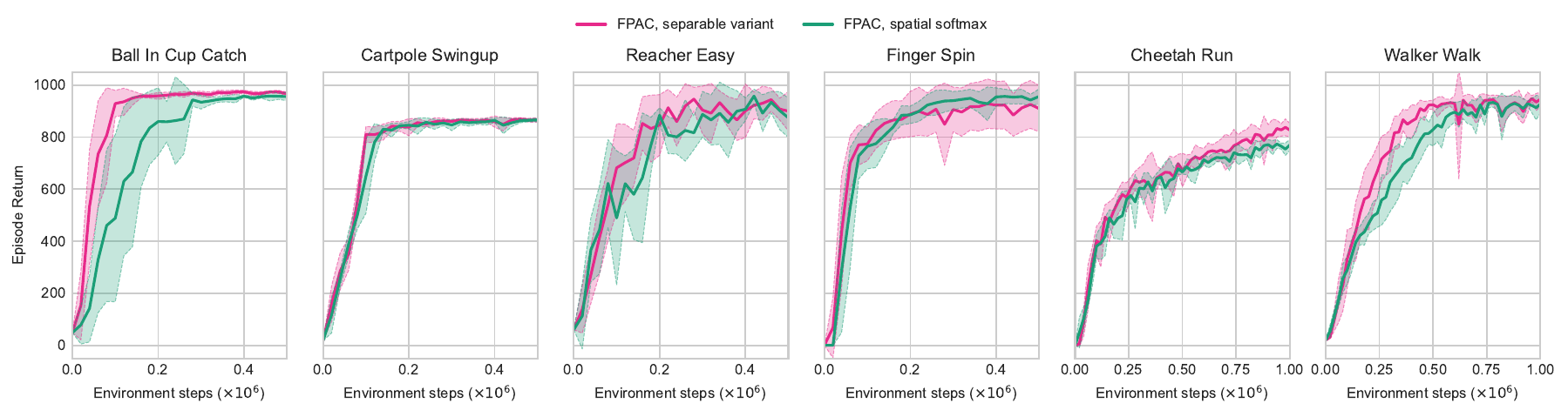}
\caption{
Impact of separable spatial softmax implementation on learning performance of FPAC.
}
\label{fig:softmax}
\end{figure*}

\begin{figure*}[t!]
\centering
\includegraphics[width=1\textwidth, trim=5 5 5 5, clip]{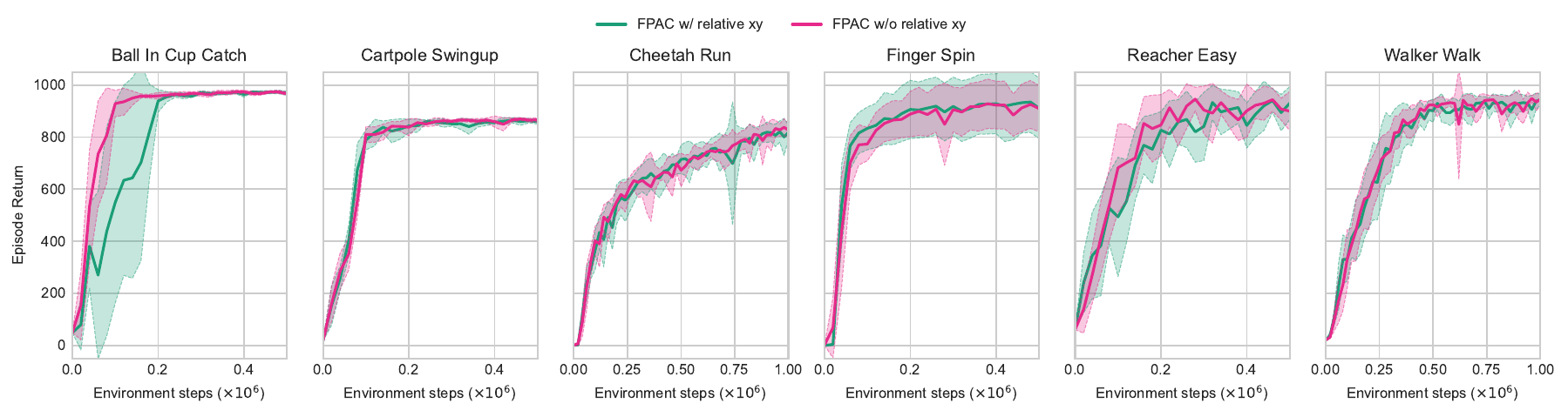}
\caption{
Impact of relative feature points on learning performance of FPAC.
}
\label{fig:relative}
\end{figure*}

% \subsection{Robustness of actor to learned feature points}

% To analyze the robustness of the policy to the learned feature points, we randomly set subsets of learned feature point coordinates and scalar features to 0 and measure the cumulative rewards of pre-trained FPAC agents. The random subset is fixed throughout an episode. The results of our experiments are shown in Fig.~\ref{fig:drop}. The performance of FPAC agents degrades smoothly as the number of dropped feature points increases and the agent is still able to reasonably perform well even after dropping 25\% of the feature points.

% \begin{figure*}[t]
% \centering
% \includegraphics[width=1\textwidth, trim=110 15 110 30, clip]{drop_keypoints.pdf}
% \caption{
% Evaluation of robustness of actor network to learned feature points. We take pre-trained FPAC agents that use 32 feature points and randomly drop a subset of (1, 2, 4, or 8) feature points to 0. We plot the mean and standard deviation of cumulative rewards obtained across 100 episodes. The performance of FPAC degrades smoothly as the number of dropped feature points increases and the agent is still able to reasonably perform the tasks even after dropping 25\% of the feature points.
% }
% \label{fig:drop}
% \end{figure*}

\subsection{Generalization of pre-trained feature points}
\label{sec:generalization}

In our experiments, we observed that pre-trained feature points perform similar to FPAC on the simpler tasks but significantly worse on Cheetah and Walker (see Fig.~\ref{fig:key}). We investigate this by looking at the predictions of the pre-trained encoder as RL training progresses. We plot some predictions in Fig.~\ref{fig:pretrain-failure}. We observe that as the RL agent learns, it visits new states that are out-of-distribution of the pre-trained encoder and the pre-trained feature points fail to generalize well on these states. This significantly decreases the data-efficiency and asymptotic performance of the RL agent. 

For further analysis, we plot the y-coordinate of the torso of the cheetah throughout a training run in Fig.~\ref{fig:pretrain-ablation}. We can observe that the distribution of this measurement is narrow in the initial random exploration phase of training. After training on more steps, exploration by the RL agent leads to upside down flipping of the cheetah. This is denoted by a jump from higher to lower y-coordinate value in Fig.~\ref{fig:pretrain-ablation}. As we show in Fig.~\ref{fig:pretrain-failure}, the pre-trained encoder does not make accurate feature point predictions in these states and subsequently the RL agent with pre-trained feature points fails to recover and continues to flip even if we train it for longer. In contrast, our FPAC agent with end-to-end training also explores flipping in the initial stages of training but quickly learns to recover (as the feature points encoder learns to extract the right feature points on newly visited states) and run forward efficiently.
% We observe that even if we continue training the pre-trained encoder on new data, it still fails to improve the performance of the RL agent.

While it might be possible to tune the pre-training so that it generalizes well, FPAC learns relevant feature points from scratch, in an end-to-end manner, to match the learning performance of SAC from handcrafted keypoints and performs competitive to the state-of-the-art methods on PlaNet and Dreamer benchmarks.

\subsection{Impact of scalar feature}

We measure the impact of scalar feature $m_k$ on the learning performance of FPAC in Fig.~\ref{fig:feature}. Use of the scalar feature has a significant impact in the Ball in Cup Catch task but only a minor impact on other tasks.

\subsection{Impact of feature point velocity term}

We measure the impact of the feature point velocity term in Fig.~\ref{fig:velocity}. We observe that while FPAC is able to learn most tasks reasonably well even without the velocity term, FPAC with the velocity term consistently performs better.

\subsection{Impact of separable spatial softmax}

The feature point coordinates can be computed as the expected values of pixel coordinates, after a spatial softmax operation \cite{levine2016end} on the convolutional feature maps \eqref{eq:key2d}. We refer to this version as \emph{FPAC, spatial softmax}. In practice, we use a separable variant that computes each coordinate separately \eqref{eq:key1d}. We refer to this version as \emph{FPAC, separable variant}. We compare the RL performance of the spatial softmax version against the separable variant in Fig.~\ref{fig:softmax}. We observe that both perform similarly and the separable variant performs slightly better than the spatial softmax version in some tasks.

The separable variant is also computationally more efficient than the spatial softmax version. We benchmark both versions on an NVIDIA GTX 1080 Ti GPU. We perform 100k calls to both functions using a batch of $128 \times 32 \times 35 \times 35$ convolutional feature maps and measure the average time required for each operation. The spatial softmax version takes 2.8 ms, with the majority of the time spent on the 2D softmax operation which takes 1.8 ms and then the 2D expectation operation for each coordinate takes 0.5 ms. The separable variant only takes 1.1 ms where the mean pooling, 1D softmax, and 1D expectation operation for each coordinate takes 0.55 ms.

% Separable softmax: 0.00111
%     - compute_coord: 0.000555
%         - mean pooling: 0.000158
%         - 1D softmax: 
%         - 1D expectation: 

% Spatial softmax: 0.00283
%     - 2D softmax: 0.00180
%     - 2D expectation: 0.000515

\subsection{Impact of relative feature points}
\label{sec:relative}

We measure the impact of using absolute vs relative feature point coordinates on the learning performance of FPAC in Fig.~\ref{fig:relative}. Use of the relative feature points only has a significant impact in the Ball in Cup Catch task. Use of absolute feature points generalizes well to all tasks.

% \subsection{Comparison of overall training time}

% We measure an overall training time of 70 minutes 41 seconds to train a SAC from pixels agent and 76 minutes 56 seconds to train our FPAC agent for 500 episodes on the Cartpole Swingup task. We report an average of 10 runs on an NVIDIA V100 GPU. The only difference between the SAC from pixels agent and our FPAC agent is the additional feature point bottleneck used in FPAC. The additional feature point bottleneck makes our approach only negligibly slower than SAC from pixels (which does not learn well).

% \section{Network architecture and hyperparameters}

% Similar to prior works, our convolutional encoder consists of four convolutional layers with 32 channels, $3 \times 3$ kernels and a stride of 2 on the first layer. These convolutional layers reduce the $3 \times 84 \times 84$ observations to $32 \times 35 \times 35$ feature maps and we additionally use a $1 \times 1$ convolutional layer to project these 32 feature maps to $K$ feature maps (to compute locations and features for $K$ feature points). For our actor and critic networks, we use MLPs with 2 hidden layers and 1024 hidden units in each layer. We use the Swish non-linearity \cite{ramachandran2017searching} in all our networks.

%===============================================================================

\section{Conclusion}
\label{sec:conclusion}

We demonstrate that it is possible to directly learn feature points that are relevant for RL from images, without any additional supervision. Our FPAC method, which only adds a simple feature points bottleneck, is easy to implement and fast to run. We demonstrate that FPAC performs competitively to the state-of-the-art methods on the PlaNet and Dreamer benchmarks. FPAC can learn feature points from scratch, even in tasks with sparse rewards, to nearly match the performance of SAC learning from low-dimensional representations. 
% We observe that FPAC learns feature points to represent parts of multiple objects. 
We observe that feature points learned end-to-end, from scratch, work at least as well as feature points learned with pre-training.
FPAC is robust to the choice of hyperparameters and can perform robustly across tasks with a different number of objects and different object dynamics. The code to reproduce our experiments is provided in the supplementary material. Potential lines of future work include: (i)~learning feature point dynamics models that can be used to generate additional data to train actor-critic networks, and (ii)~learning 3D feature points from monocular images or multiple views. %, and (iii)~applying FPAC to real-world robotic manipulation tasks.

%===============================================================================

% \section*{Acknowledgment}

\bibliographystyle{IEEEtran}
\bibliography{IEEEabrv, references}

\begin{thebibliography}{10}
\providecommand{\url}[1]{#1}
\csname url@samestyle\endcsname
\providecommand{\newblock}{\relax}
\providecommand{\bibinfo}[2]{#2}
\providecommand{\BIBentrySTDinterwordspacing}{\spaceskip=0pt\relax}
\providecommand{\BIBentryALTinterwordstretchfactor}{4}
\providecommand{\BIBentryALTinterwordspacing}{\spaceskip=\fontdimen2\font plus
\BIBentryALTinterwordstretchfactor\fontdimen3\font minus
  \fontdimen4\font\relax}
\providecommand{\BIBforeignlanguage}[2]{{%
\expandafter\ifx\csname l@#1\endcsname\relax
\typeout{** WARNING: IEEEtran.bst: No hyphenation pattern has been}%
\typeout{** loaded for the language `#1'. Using the pattern for}%
\typeout{** the default language instead.}%
\else
\language=\csname l@#1\endcsname
\fi
#2}}
\providecommand{\BIBdecl}{\relax}
\BIBdecl

\bibitem{tassa2020dmcontrol}
Y.~Tassa, S.~Tunyasuvunakool, A.~Muldal, Y.~Doron, P.~Trochim, S.~Liu,
  S.~Bohez, J.~Merel, T.~Erez, T.~Lillicrap \emph{et~al.}, ``dm\_control:
  Software and tasks for continuous control,'' \emph{arXiv preprint
  arXiv:2006.12983}, 2020.

\bibitem{yarats2019improving}
D.~Yarats, A.~Zhang, I.~Kostrikov, B.~Amos, J.~Pineau, and R.~Fergus,
  ``Improving sample efficiency in model-free reinforcement learning from
  images,'' \emph{arXiv preprint arXiv:1910.01741}, 2019.

\bibitem{laskin_srinivas2020curl}
M.~Laskin, A.~Srinivas, and P.~Abbeel, ``{CURL}: Contrastive unsupervised
  representations for reinforcement learning,'' in \emph{International
  Conference on Machine Learning}, 2020.

\bibitem{laskin2020reinforcement}
M.~Laskin, K.~Lee, A.~Stooke, L.~Pinto, P.~Abbeel, and A.~Srinivas,
  ``Reinforcement learning with augmented data,'' in \emph{Advances in Neural
  Information Processing Systems}, vol.~33, 2020.

\bibitem{yarats2021image}
D.~Yarats, I.~Kostrikov, and R.~Fergus, ``Image augmentation is all you need:
  Regularizing deep reinforcement learning from pixels,'' in
  \emph{International Conference on Learning Representations}, 2021.

\bibitem{haarnoja2018softa}
T.~Haarnoja, A.~Zhou, P.~Abbeel, and S.~Levine, ``Soft actor-critic: Off-policy
  maximum entropy deep reinforcement learning with a stochastic actor,'' in
  \emph{International Conference on Machine Learning}, 2018, pp. 1861--1870.

\bibitem{haarnoja2018softb}
T.~Haarnoja, A.~Zhou, K.~Hartikainen, G.~Tucker, S.~Ha, J.~Tan, V.~Kumar,
  H.~Zhu, A.~Gupta, P.~Abbeel \emph{et~al.}, ``Soft actor-critic algorithms and
  applications,'' \emph{arXiv preprint arXiv:1812.05905}, 2018.

\bibitem{levine2016end}
S.~Levine, C.~Finn, T.~Darrell, and P.~Abbeel, ``End-to-end training of deep
  visuomotor policies,'' \emph{The Journal of Machine Learning Research},
  vol.~17, no.~1, pp. 1334--1373, 2016.

\bibitem{finn2016deep}
C.~Finn, X.~Y. Tan, Y.~Duan, T.~Darrell, S.~Levine, and P.~Abbeel, ``Deep
  spatial autoencoders for visuomotor learning,'' in \emph{2016 IEEE
  International Conference on Robotics and Automation (ICRA)}.\hskip 1em plus
  0.5em minus 0.4em\relax IEEE, 2016, pp. 512--519.

\bibitem{cabi2019scaling}
S.~Cabi, S.~G. Colmenarejo, A.~Novikov, K.~Konyushkova, S.~Reed, R.~Jeong,
  K.~Zolna, Y.~Aytar, D.~Budden, M.~Vecerik \emph{et~al.}, ``Scaling
  data-driven robotics with reward sketching and batch reinforcement
  learning,'' \emph{Robotics: Science and Systems}, 2020.

\bibitem{jakab2018unsupervised}
T.~Jakab, A.~Gupta, H.~Bilen, and A.~Vedaldi, ``Unsupervised learning of object
  landmarks through conditional image generation,'' in \emph{Advances in Neural
  Information Processing Systems}, vol.~31, 2018.

\bibitem{zhang2018unsupervised}
Y.~Zhang, Y.~Guo, Y.~Jin, Y.~Luo, Z.~He, and H.~Lee, ``Unsupervised discovery
  of object landmarks as structural representations,'' in \emph{IEEE Conference
  on Computer Vision and Pattern Recognition}, 2018, pp. 2694--2703.

\bibitem{kulkarni2019unsupervised}
T.~D. Kulkarni, A.~Gupta, C.~Ionescu, S.~Borgeaud, M.~Reynolds, A.~Zisserman,
  and V.~Mnih, ``Unsupervised learning of object keypoints for perception and
  control,'' in \emph{Advances in Neural Information Processing Systems},
  vol.~32, 2019, pp. 10\,724--10\,734.

\bibitem{gopalakrishnan2020unsupervised}
A.~Gopalakrishnan, S.~van Steenkiste, and J.~Schmidhuber, ``Unsupervised object
  keypoint learning using local spatial predictability,'' in
  \emph{International Conference on Learning Representations}, 2021.

\bibitem{minderer2019unsupervised}
M.~Minderer, C.~Sun, R.~Villegas, F.~Cole, K.~Murphy, and H.~Lee,
  ``Unsupervised learning of object structure and dynamics from videos,'' in
  \emph{Advances in Neural Information Processing Systems}, 2019.

\bibitem{manuelli2020keypoints}
L.~Manuelli, Y.~Li, P.~Florence, and R.~Tedrake, ``Keypoints into the future:
  Self-supervised correspondence in model-based reinforcement learning,'' in
  \emph{Conference on Robot Learning}, 2020.

\bibitem{thewlis2017unsupervised}
J.~Thewlis, H.~Bilen, and A.~Vedaldi, ``Unsupervised learning of object
  landmarks by factorized spatial embeddings,'' in \emph{IEEE International
  Conference on Computer Vision}, 2017, pp. 5916--5925.

\bibitem{thewlis2019unsupervised}
J.~Thewlis, S.~Albanie, H.~Bilen, and A.~Vedaldi, ``Unsupervised learning of
  landmarks by descriptor vector exchange,'' in \emph{IEEE International
  Conference on Computer Vision}, 2019, pp. 6361--6371.

\bibitem{lange2010deep}
S.~Lange and M.~A. Riedmiller, ``Deep learning of visual control policies,'' in
  \emph{ESANN}, 2010.

\bibitem{shelhamer2016loss}
E.~Shelhamer, P.~Mahmoudieh, M.~Argus, and T.~Darrell, ``Loss is its own
  reward: Self-supervision for reinforcement learning,'' \emph{arXiv preprint
  arXiv:1612.07307}, 2016.

\bibitem{higgins2017darla}
I.~Higgins, A.~Pal, A.~Rusu, L.~Matthey, C.~Burgess, A.~Pritzel, M.~Botvinick,
  C.~Blundell, and A.~Lerchner, ``Darla: Improving zero-shot transfer in
  reinforcement learning,'' in \emph{International Conference on Machine
  Learning}, 2017, pp. 1480--1490.

\bibitem{nair2018visual}
A.~V. Nair, V.~Pong, M.~Dalal, S.~Bahl, S.~Lin, and S.~Levine, ``Visual
  reinforcement learning with imagined goals,'' in \emph{Advances in Neural
  Information Processing Systems}, 2018, pp. 9191--9200.

\bibitem{hafner2019learning}
D.~Hafner, T.~Lillicrap, I.~Fischer, R.~Villegas, D.~Ha, H.~Lee, and
  J.~Davidson, ``Learning latent dynamics for planning from pixels,'' in
  \emph{International Conference on Machine Learning}, 2019, pp. 2555--2565.

\bibitem{hafner2019dream}
D.~Hafner, T.~Lillicrap, J.~Ba, and M.~Norouzi, ``Dream to control: Learning
  behaviors by latent imagination,'' in \emph{International Conference on
  Learning Representations}, 2019.

\bibitem{ha2018recurrent}
D.~Ha and J.~Schmidhuber, ``Recurrent world models facilitate policy
  evolution,'' in \emph{Advances in Neural Information Processing Systems},
  2018, pp. 2455--2467.

\bibitem{singh2019}
A.~Singh, L.~Yang, K.~Hartikainen, C.~Finn, and S.~Levine, ``End-to-end robotic
  reinforcement learning without reward engineering,'' \emph{Robotics: Science
  and Systems}, 2019.

\bibitem{zhu2019ingredients}
H.~Zhu, J.~Yu, A.~Gupta, D.~Shah, K.~Hartikainen, A.~Singh, V.~Kumar, and
  S.~Levine, ``The ingredients of real world robotic reinforcement learning,''
  in \emph{International Conference on Learning Representations}, 2019.

\bibitem{kendall2019learning}
A.~Kendall, J.~Hawke, D.~Janz, P.~Mazur, D.~Reda, J.-M. Allen, V.-D. Lam,
  A.~Bewley, and A.~Shah, ``Learning to drive in a day,'' in \emph{IEEE
  International Conference on Robotics and Automation}, 2019, pp. 8248--8254.

\bibitem{viitala2020learning}
A.~Viitala, R.~Boney, Y.~Zhao, A.~Ilin, and J.~Kannala, ``Learning to drive
  ({L2D}) as a low-cost benchmark for real-world reinforcement learning,''
  \emph{arXiv preprint arXiv:2008.00715}, 2020.

\bibitem{lee2020stochastic}
A.~Lee, A.~Nagabandi, P.~Abbeel, and S.~Levine, ``Stochastic latent
  actor-critic: Deep reinforcement learning with a latent variable model,'' in
  \emph{Advances in Neural Information Processing Systems}, vol.~33, 2020.

\end{thebibliography}

\end{document}